\documentclass[journal,twoside]{IEEEtran}
\usepackage[ruled,linesnumbered]{algorithm2e}

\SetAlFnt{\small}
\SetAlCapFnt{\small}
\SetAlCapNameFnt{\small}
\SetAlCapHSkip{0pt}
\IncMargin{-\parindent}

\usepackage{cite}
\usepackage{bbm}
\usepackage{array}
\usepackage{mathrsfs}
\usepackage{makecell}
\usepackage{etoolbox}
\usepackage{multirow}
\usepackage[flushleft]{threeparttable}
\usepackage{url,subfigure,amsmath,amssymb,epsfig,verbatim,booktabs,graphicx,epstopdf}
\usepackage[colorlinks=false,linkcolor=black, citecolor=blue, urlcolor=black, pdfborder={0 1 0}]{hyperref}
\usepackage{mathtools}
\usepackage[justification=centering]{caption}
\usepackage{xcolor}

\begin{document}
%\begin{frontmatter}
\title{Deep Reinforcement Learning for Solving the Heterogeneous Capacitated Vehicle Routing Problem}
\author{Jingwen Li, Yining Ma, Ruize Gao, Zhiguang Cao$^\dagger$, Andrew Lim, Wen Song, Jie Zhang \\% <-this % stops a space

\thanks{$^\dagger$ Corresponding author.}
\thanks{Jingwen Li and Yining Ma are with the Department of Industrial Systems Engineering and Management, National University of Singapore, Singapore (emails: lijingwen@u.nus.edu, yiningma@u.nus.edu).}
\thanks{Ruize Gao is with the Departmant of Computer Science and Engineering, Chinese Universiry of Hong Kong, China (email: ruizegao@cuhk.edu.hk).}
\thanks{Zhiguang Cao is with the Singapore Institute of Manufacturing Technology, Singapore (email: zhiguangcao@outlook.com).}
\thanks{Andrew Lim is with the School of Computing and Artificial Intelligence, Southwest Jiaotong University, China (email: i@limandrew.org).}% <-this % stops a space
\thanks{Wen Song is with the Institute of Marine Science and Technology, Shandong University, China (email: wensong@email.sdu.edu.cn).}% <-this % stops a space
\thanks{Jie Zhang is with the School of Computer Science and Engineering, Nanyang Technological University, Singapore (email: zhangj@ntu.edu.sg).}

}

%The paper headers
\markboth{
IEEE TRANSACTIONS ON CYBERNETICS, 2021}%
{Li \MakeLowercase{\textit{et al.}}: Deep Reinforcement Learning for Solving the Heterogeneous Capacitated Vehicle Routing Problem}

\maketitle
\begin{abstract}
Existing deep reinforcement learning (DRL) based methods for solving the capacitated vehicle routing problem (CVRP) intrinsically cope with homogeneous vehicle fleet, in which the fleet is assumed as repetitions of a single vehicle. Hence, their key to construct a solution solely lies in the selection of the next node (customer) to visit excluding the selection of vehicle. However, vehicles in real-world scenarios are likely to be heterogeneous with different characteristics that affect their capacity (or travel speed), rendering existing DRL methods less effective. In this paper, we tackle heterogeneous CVRP (HCVRP), where vehicles are mainly characterized by different capacities. We consider both min-max and min-sum objectives for HCVRP, which aim to minimize the longest or total travel time of the vehicle(s) in the fleet. To solve those problems, we propose a DRL method based on the attention mechanism with a vehicle selection decoder accounting for the heterogeneous fleet constraint and a node selection decoder accounting for the route construction, which learns to construct a solution by automatically selecting both a vehicle and a node for this vehicle at each step. Experimental results based on randomly generated instances show that, with desirable generalization to various problem sizes, our method outperforms the state-of-the-art DRL method and most of the conventional heuristics, and also delivers competitive performance against the state-of-the-art heuristic method, i.e., SISR. Additionally, the results of extended experiments demonstrate that our method is also able to solve CVRPLib instances with satisfactory performance.

\end{abstract}


%\end{frontmatter}
%--------------------------------------------------------------------Introduction
\section{Introduction}
\label{sec:intro}

\IEEEPARstart{T}{he} Capacitated Vehicle Routing Problem (CVRP) is a classical combinatorial optimization problem, which aims to optimize the routes for a fleet of vehicles with capacity constraints to serve a set of customers with demands. Compared with the assumption of multiple identical vehicles in homogeneous CVRP, the settings of vehicles with different capacities (or speeds) are more in line with the real-world practice, which leads to the heterogeneous CVRP (HCVRP)~\cite{golden1984fleet,kocc2015hybrid}. According to the objectives, CVRP can also be classified as min-max and min-sum ones, respectively. The former objective requires that the longest (worst-case) travel time (or distance) for a vehicle in the fleet should be as satisfying as possible since fairness is crucial in many real-world applications\cite{wang2016min,duran2011pre,bertazzi2015min,ma2010min,alabas2008self,hashi2016gis,liu2006optimization}, and the latter objective aims to minimize the total travel time (or distance) incurred by the whole fleet~\cite{haimovich1985bounds,lysgaard2004new,szeto2011artificial,wang2017vehicle}. In this paper, we study the problem of HCVRP with both min-max and min-sum objectives, i.e., MM-HCVRP and MS-HCVRP.

Conventional methods for solving HCVRP include exact and heuristic ones. Exact methods usually adopt branch-and-bound or its variants as the framework and perform well on small-scale problems~\cite{baldacci2009unified,lysgaard2004new,francca1995m,haimovich1985bounds}, but may consume prohibitively long time on large-scale ones given the exponential computation complexity. Heuristic methods usually exploit certain hand-engineered searching rules to guide the solving processes, which often consume much shorter time and are more desirable for large-scale 
problems in reality~\cite{mostafa2017solving,li2010adaptive,szeto2011artificial,yakici2017heuristic}. However, such hand-engineered rules largely rely on human experience and domain knowledge, thus might be incapable of engendering solutions with high quality. 
Moreover, both conventional exact and heuristic methods always solve the problem instances independently, and fail to exploit the patterns that potentially shared among the instances.

Recently, researchers tend to apply deep reinforcement learning (DRL) to automatically learn the searching rules in heuristic methods for solving routing problems including CVRP and TSP~\cite{bello2017neural,xin2021multi,nazari2018reinforcement,kool2018attention,chen2019learning,li2021heterogeneous}, by discovering the underlying patterns from a large number of instances. Generally, those DRL models are categorized as two classes, i.e., construction and improvement methods, respectively. Starting with an empty solution, the former constructs a solution by sequentially assigning each customer to a vehicle until all customers are served. Starting with a complete initial solution, the latter selects either candidate nodes (customers or depot) or heuristic operators, or both to improve and update the solution at each step, which are repeated until termination. By further leveraging the advanced deep learning architectures like attention mechanism to guide the selection,
those DRL models are able to efficiently generate solutions with much higher quality compared to conventional heuristics. However, existing works only focus on solving homogeneous CVRP which intrinsically cope with vehicles of the same characteristics, in the sense that the complete route of the fleet could be derived by repeatedly dispatching a single vehicle. Consequently, the key in those works is to solely select the next node to visit excluding the selection of vehicles, since there is only one vehicle essentially. Evidently, those works would be far less effective when applied to solve the more practical HCVRP, given the following issues: 1) The assumption of homogeneous vehicles is unable to capture the discrepancy in vehicles; 2) The vehicle selection is not explicitly considered, which should be of equal importance to the node selection in HCVRP; 3) The contextual information in the attention scheme is insufficient as it lacks states of other vehicles and (partially) constructed routes, which may render it incapable of engendering high-quality solutions in view of the complexity of HCVRP.

In this paper, we aim to solve the HCVRP with both min-sum and min-max objectives while emphasizing on addressing the aforementioned issues. We propose a novel neural architecture integrated with the attention mechanism to improve the DRL based construction method, which combines the decision-making for vehicle selection and node selection together to engender solutions of higher quality. Different from the existing works that construct the routes for each vehicle of the homogeneous fleet in sequence, our policy network is able to automatically and flexibly select a vehicle from a heterogeneous fleet at each step. In specific, our policy network adopts a Transformer-style \cite{vaswani2017attention} encoder-decoder structure, where the decoder consists of two parts, i.e., vehicle selection decoder and node selection decoder, respectively. 
With the problem features (i.e., customer location, customer demand and vehicle capacity) processed by the encoder for better representation, the policy network first selects a vehicle from the fleet using the vehicle selection decoder based on the states of all vehicles and partial routes, and then selects a node for this vehicle using the node selection decoder at each decoding step. This process is repeated until all customers are served.

Accordingly, the major contribution of this paper is that we present a deep reinforcement learning method to solve CVRP with multiple heterogeneous vehicles, which is intrinsically different from the homogeneous ones in existing works, as the latter is lacking in selecting vehicles from a fleet. Specifically, we propose an effective neural architecture that integrates the vehicle selection and node selection together, with rich contextual information for selection among the heterogeneous vehicles, where every vehicle in the fleet has the chance to be selected at each step. We test both min-max and min-sum objectives with various sizes of vehicles and customers. Results show that our method is superior to most of the conventional heuristics and competitive to the state-of-the-art heuristic (i.e., SISR) with much shorter computation time. With comparable computation time, our method achieves much better solution quality than that of other DRL method. In addition, our method generalizes well to problems with larger customer sizes.

The remainder of the paper is organized as follows. Section \ref{sec:related work} briefly reviews conventional methods and deep models for routing problems. Section \ref{sec:problem descriptiton} introduces the mathematical formulation of MM-HCVRP and MS-HCVRP and the reformulation in the RL (reinforcement learning) manner. Section \ref{sec:method} elaborates our DRL framework. Section \ref{sec:experiments} provides the computational experiments and analysis. Finally, Section \ref{sec:conclusion} concludes the paper and presents future works. 

%-------------------------------------------------------------------------------Reinforcement
\section{Related Works}
\label{sec:related work}
In this section, we briefly review the conventional methods for solving HCVRP with different objective functions, and deep models for solving the general VRPs. 

The heterogeneous CVRP (HCVRP) was first studied in~\cite{golden1984fleet}, where the Clarke and Wright procedure and partition algorithms were applied to generate the lower bound and estimate optimal solution. An efficient constructive heuristic was adopted to solve HCVRP in~\cite{prins2002efficient} by merging small start trips for each customer into a complete one, which was also capable for multi-trip cases. 
Baldacci and Mingozzi~\cite{baldacci2009unified} presented a unified exact method to solve HCVRP, reducing the number of variables by using three bounding procedures. 
Feng et al.~\cite{feng2019solving} proposed a novel evolutionary multitasking algorithm to tackle the HCVRP with time window, and occasional driver, which can also solve multiple optimization tasks simultaneously. 

The CVRP with min-sum objective was first proposed by Dantzig and Ramsey~\cite{dantzig1959truck}, which was assumed as the generalization of Travelling Salesman Problem (TSP) with capacity constraints. 
To address the large-scale multi-objective optimization problem (MOP), a competitive swarm optimizer (CSO) based search method was proposed in~\cite{tian2019efficient, cheng2014competitive, cheng2016test}, which conceived a new particle updating strategy to improve the search accuracy and efficiency. By transforming the large-scale CVRP (LSCVRP) into a large-scale MOP, an evolutionary multi-objective route grouping method was introduced in~\cite{xiao2019evolutionary}, which employed a multi-objective evolutionary algorithm to decompose the LSCVRP into small tractable sub-components.
%, after which a local search method was applied to improve the solutions.
The min-max objective was considered in a multiple Travelling Salesman Problem (TSP)~\cite{francca1995m}, which was solved by a tabu search heuristic and two exact search schemes. 
An ant colony optimization method was proposed to address the min-max Single Depot CVRP (SDCVRP)~\cite{narasimha2011ant}. 
The problem was further extended to the min-max multi-depot CVRP~\cite{narasimha2013ant}, which could be reduced to SDCVRP using an equitable region partitioning approach. 
A swarm intelligence based heuristic algorithm was presented to address the rich min-max CVRP~\cite{yakici2017heuristic}. 
The min-max cumulative capacitated vehicle routing problem, aiming to minimize the last arrival time at customers, was first studied in~\cite{sze2017cumulative,golden1997adaptive}, where a two-stage adaptive variable neighbourhood search (AVNS) algorithm was introduced and also tested in min-sum objective to verify generalization.

The first deep model for routing problems is Pointer Network, which used supervised learning to solve TSP \cite{vinyals2015pointer} and was later extended to reinforcement learning  \cite{bello2017neural}. Afterwards, Pointer Network was adopted to solve CVRP in \cite{nazari2018reinforcement}, where the Recurrence Neural Network architecture in the encoder was removed to reduce computation complexity without degrading solution quality. To further improve the performance, a Transformer based architecture was incorporated by integrating self-attention in both the encoder and decoder~\cite{kool2018attention}. Different from the above methods which learn constructive heuristics, NeuRewriter was proposed to learn how to pick the next solution in a local search framework~\cite{chen2019learning}. 
Despite their promising results, these methods are less effective for tackling the heterogeneous fleet in HCVRP. Recently, some learning based methods have been proposed to solve HCVRP. Inspired by multi-agent RL, Vera and Abad \cite{vera2019deep} made the first attempt to solve the min-sum HCVRP through cooperative actions of multiple agents for route construction. Qin et al. \cite{qin2021novel} proposed a reinforcement learning based controller to select among several meta-heuristics with different characteristics to solve min-sum HCVRP. Although yielding better performance than conventional heuristics, they are unable to well handle either the min-max objective or heterogeneous speed of vehicles.

% \begin{figure*}%[!htbp] 
% \centering 
%      \setlength{\abovecaptionskip}{0.1cm}
%  	\setlength{\belowcaptionskip}{-0.5cm}
% 	\includegraphics[width=0.96\textwidth, height=58mm, trim={1mm 0mm 1mm 0mm},clip]{figs/MDP.png} 
% 	\captionsetup{justification=justified}
% 	\caption{
% 	An illustrative MDP with 2 vehicles and 4 customers. At the first step, vehicle $v^1$ is selected to serve customer $x^1$. In response to this action: a) the remaining capacity of vehicle $v^1$ is updated by subtracting demand of customer $x^1$, i.e., $o_1^1=o_0^1-d_0^1$ and the demand of customer $x^1$ is set to zero from then on, i.e., $d^1_t = 0$ for $t \geq 1 $; b) the travel time between node $x^0$ and $x^1$ is added to the accumulate travel time of vehicle $v^1$, i.e., $T_1^1 = T_0^1  + D(x^0,x^1)/f$; and c) the partial routes of all vehicles are updated, i.e., $G^1_1 = \{ 0,1\}$ and $G^2_1 = \{ 0,0\}$. At the second step, vehicle $v^2$ is then selected to serve customer $x^4$, and all the relevant variables are updated similarly. 
% 	}
% 	\label{fig:MDP} 
% \end{figure*}

\section{Problem Formulation}
\label{sec:problem descriptiton} 
In this section, we first introduce the mathematical formulation of HCVRP with both min-max and min-sum objectives, and then reformulate it as the form of reinforcement learning.

\subsection{Mathematical Formulation of HCVRP}
Particularly, with $n+1$ nodes (customers and depot) represented as $ X=\{ x^i \}_{i=0}^n $ and node $x^0$ denoting the depot, the customer set is assumed to be $X' = X\setminus \{x^0\}$. Each node $x^i \in \mathbb{R}^3$ is defined as $\{ (s^i, d^i)\}$, where $s^i$ contains the 2-dim location coordinates of node $x^i$, and $d^i$ refers to its demand (the demand for depot is 0). Here, we take heterogeneous vehicles with different capacities into account, which respects the real-world situations. Accordingly, let $ V=\{ v^i \}_{i=1}^m$ represent the heterogeneous fleet of vehicles, where each element $v^i$ is defined as $\{(\mathcal{Q}^i) \}$, i.e., the capacity of vehicle $v^i$. 
The HCVRP problem describes a process that all fully loaded vehicles start from depot, and sequentially visit the locations of customers to satisfy their demands, with the constraints that each customer can be visited exactly once, and the loading amount for a vehicle during a single trip can never exceed its capacity. 

Let $D(x^i,x^j)$ be the Euclidean distance between $x^i$ and $x^j$. Let $y_{ij}^v$ be a binary variable, which equals to 1 if vehicle $v$ travels directly from customer $x^i$ to $x^j$, and 0 otherwise. Let $l_{ij}^v$ be the remaining capacity of the vehicle $v$ before travelling from customer $x^i$ to customer $x^j$. For simplification, we assume that all vehicles have the same speed $f$, which could be easily extended to take different values. Then, the MM-HCVRP could be naturally defined as follows,
\begin{equation}
     \min\ \max_{v\in V}(\sum_{i \in X} \sum_{j \in X} \frac{D(x^i,x^j)}{f} y_{ij}^v).
\end{equation}
subject to the following six constraints, 
\begin{align}
\sum_{v \in V}\sum_{j \in X} y_{ij}^v = 1,\qquad i \in X' 
\label{constraint1}
\end{align}
\begin{align}
\sum_{i \in X} y_{ij}^v - \sum_{k \in X} y_{jk}^v = 0,\qquad v\in V, j \in X'
\label{constraint2}
\end{align}
\begin{align}
\sum_{v \in V} \sum_{i \in X} l_{ij}^v - \sum_{v \in V} \sum_{k \in X} l_{jk}^v = d^j,\qquad j \in X'
\label{constraint3}
\end{align}
\begin{align}
d^j y_{ij}^v\leq l_{ij}^v \leq  \left(\mathcal{Q}^v - d^i\right) \cdot y_{ij}^v,\qquad v\in V,i\in X, j \in X
\label{constraint4}
\end{align}
\begin{align}
y_{ij}^v = \left\{ 0,1 \right\},\qquad v\in V,i\in X, j \in X
\label{constraint5}
\end{align}
\begin{align}
l_{ij}^v \geq 0,\ d^i \geq 0,  \qquad v\in V,i\in X, j \in X.
\label{constraint6}
\end{align}

The objective of the formulation is to minimize the maximum travel time among all vehicles. Constraint (\ref{constraint1}) and (\ref{constraint2}) ensure that each customer is visited exactly once and each route is completed by the same vehicle. Constraint (\ref{constraint3}) guarantees that the difference between amount of goods loaded by a vehicle before and after serving a customer equals to the demand of that customer. Constraint (\ref{constraint4}) enforces that the amount of goods for any vehicle is able to meet the demands of the corresponding customers and never exceed its capacity. Constraint (\ref{constraint5}) defines the binary variable and constraint (\ref{constraint6}) imposes the non-negativity of the variables.

The MS-HCVRP shares the same constraints with MM-HCVRP, while the objective is formulated as follows,
\begin{equation}
    \text{min} \sum_{v\in V}\sum_{i \in X} \sum_{j \in X}  \frac{D(x^i, x^j)}{f^v} y_{ij}^v,
\end{equation}
where $f^v$ represents the speed of vehicle $v$, and it may vary with different vehicles. Thereby, it is actually minimizing the total travel time incurred by the whole fleet. 

\subsection{Reformulation as RL Form}
\label{RL_form}
Reinforcement learning (RL) was originally proposed for sequential decision-making problems, such as self-driving cars, robotics, games, etc~\cite{liu2015reinforcement,modares2015optimized,wang2016reinforcement,bai2019adaptive,wen2019online,nguyen2020deep}. The construction of routes for HCVRP step by step can be also deemed as a sequential decision-making problem. In our work, we model such process as a Markov Decision Process (MDP) \cite{bellman1957markovian} defined by 4-tuple $M=\{S,A,\tau, r\}$ (An example of the MDP is illustrated in the supplementary material).
% where an example of the MDP is illustrated in Fig.~\ref{fig:MDP}. 
Meanwhile, the detailed definition of the state space $S$, the action space $A$, the state transition rule $\tau$, and the reward function $r$ are introduced as follows. 

\textbf{State:} In our MDP, each state $s_t\!=\!(V_t, X_t)\!\in\!S$ is composed of two parts. The first part is the vehicle state $V_t$, which is expressed as $V_t\!=\!\{v_t^1, v_t^2, .... v_t^m \}\!=\!\{(o_t^1, T_t^1, G_t^1), (o_t^2, T_t^2, G_t^2), ..., (o_t^m, T_t^m, G_t^m) \}$, where  $o_t^i$ and $T_t^i$ represent the remaining capacity and the accumulate travel time of the vehicle $v^i$ at step $t$, respectively.  $G_t^i=\{g_0^i, g_1^i, ..., g_t^i \}$ represents the partial route of the vehicle $v^i$ at step $t$, where $g_j^i$ refers to the node visited by the vehicle $v^i$ at step $j$. Note that the dimension of partial routes (the number of nodes in a route) for all vehicles keeps the same, i.e., if the vehicle $v^i$ is selected to serve the node $x^j$ at step $t$, other vehicles still select their last served nodes.
Upon departure from the depot (i.e., $t=0$), the initial vehicle state is set to $V_0 = \{(\mathcal{Q}^1, 0, \{0\}), (\mathcal{Q}^2, 0, \{0\}), ..., (\mathcal{Q}^m, 0, \{0\}) \}$ where $\mathcal{Q}^i$ is the maximum capacity of vehicle $v^i$. 
The second part is the node state $S_t$, which is expressed as $X_t=\{x_t^0, x_t^1, ..., x_t^n \}=\{(s^0, d_t^0), (s^1, d_t^1), ..., (s^n, d_t^n) \}$, where $s^i$ is a 2-dim vector representing the locations of the node, and $d_t^i$ is a scalar representing the demand of node $i$ ($d_t^i$ will become 0 once that node has been served). Here, we do not consider demand splitting, and only nodes with $d^i >0$ need to be served.

\textbf{Action:} The action in our method is defined as selecting a vehicle and a node (a customer or the depot) to visit. In specific, the action $a_t \in A$ is represented as $(v_t^i, x_t^j)$, i.e., the selected node $x^j$ will be served (or visited) by the vehicle $v^i$ at step $t$. Note that only one vehicle is selected at each step.

\textbf{Transition:} The transition rule $\tau$ will transit the previous state $s_t$ to the next state $s_{t+1}$ based on the performed action $a_t = (v_t^i, x_t^j)$,  i.e., $s_{t+1} = (V_{t+1}, X_{t+1}) = \tau(V_{t}, X_{t})$.
The elements in vehicle state $V_{t+1}$ are updated as follows,
\begin{equation}
o_{t+1}^k = \left\{\begin{matrix}
o_t^k - d_t^j, & \text{if } k=i,\\
o_t^k, & \text{otherwise},
\end{matrix}\right.
\end{equation}
\begin{equation}
T_{t+1}^k = \left\{\begin{matrix}
T_t^k + \frac{D({g_t^k}, x^j)}{f}, & \text{if } k=i,\\
T_t^k, & \text{otherwise},
\end{matrix}\right.
\end{equation}
\begin{equation}
G_{t+1}^k = \left\{\begin{matrix}
[G_t^k, x^j], & \text{if } k=i,\\
[G_t^k, g_t^k], & \text{otherwise},
\end{matrix}\right.
\end{equation}
where $g_t^k$ is the last element in $G _t^k$, i.e., last visited customer by vehicle $v^k$ at step $t$, and $[\cdot, \cdot, \cdot]$ is the concatenation operator.
The element in node state $X_{t+1}$ is updated as follows,
\begin{equation}
d_{t+1}^l = \left\{\begin{matrix}
0, & \text{if } l=j,\\
d_t^l, & \text{otherwise},
\end{matrix}\right.
\end{equation}
where each demand will retain 0 after being visited.

\textbf{Reward:} For the MM-HCVRP, to minimize the maximum travel time of all vehicles, the reward is defined as the negative value of this maximum, where the reward is calculated by accumulating the travel time of multiple trips for each vehicle, respectively. Accordingly, the reward is represented as $R = - \max_{v\in V} \{ \sum_{t=0}^T \boldsymbol{r}_t\}$, where $\boldsymbol{r}_t$ is the incremental travel time for all vehicles at step $t$. Similarly, for the MS-HCVRP, the reward is defined as the negative value of the total travel time of all vehicles, i.e., $R=-\sum_{i=1}^m \sum_{t=1}^T \boldsymbol{r}_t$.
Particularly, assume that node $x^j$ and $x^k$ are selected at step $t$ and $t+1$, respectively, which are both served by the vehicle $v^i$, then the reward $\boldsymbol{r}_{t+1}$ is expressed as a $m$-dim vector as follows,
\begin{equation}
\begin{split}
\boldsymbol{r}_{t+1} = r(s_{t+1}, a_{t+1}) &= r((V_{t+1}, X_{t+1}), (v_{t+1}^i, x_{t+1}^k)) \\
&=\! \{ 0,...,0, D(\!x^j, \!x^k\!)/f,0,..., 0 \},
\end{split}
\end{equation}
where $D(x^j, x^k)/f$ is the time consumed by the vehicle $v^i$ for traveling from node $x^j$ to $x^k$, with other elements in $r(s_{t+1}, a_{t+1})$ equal to 0.

\section{Methodology}
\label{sec:method}

In this section, we introduce our deep reinforcement learning (DRL) based approach for solving HCVRP with both min-max and min-sum objectives. We first propose a novel attention-based deep neural network to represent the policy, which enables both vehicle selection and node selection at each decision step. Then we describe the procedure for training our policy network.
\begin{figure}[t] 
\centering 
 	\setlength{\belowcaptionskip}{-0.5cm}
	\includegraphics[width=0.38\textwidth,height=58mm, trim=0 5 0 20]{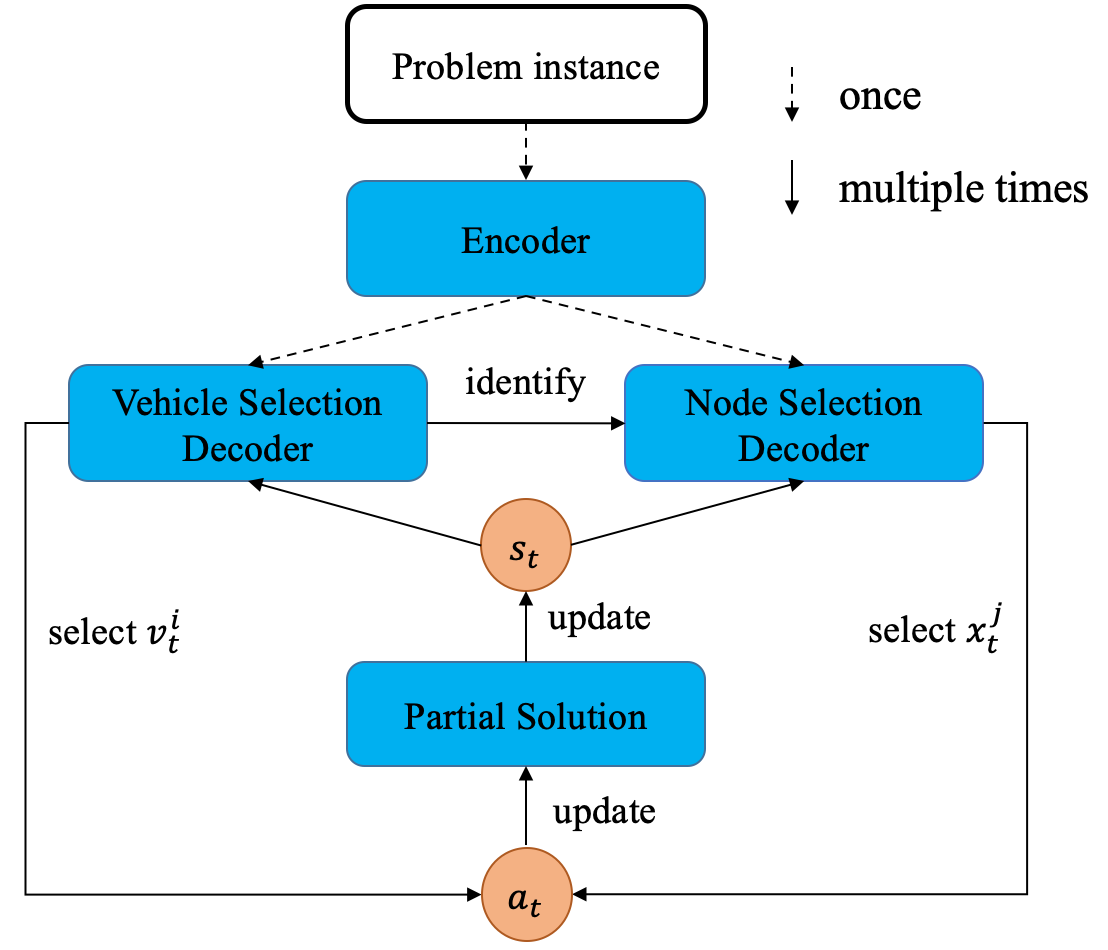} 
	\captionsetup{justification=justified}
	\caption{The framework of our policy network. With raw features of the instance processed by the encoder, our policy network first selects a vehicle ($v_t^i$) using the vehicle selection decoder and then a node ($x_t^j$) using the node selection decoder for this vehicle to visit at each route construction step $t$.  Both the selected vehicle and node constitute the action at that step, i.e., $a_t=(v_t^i, x_t^j)$, where the partial solution and state are updated accordingly. To a single instance, the encoder is executed once, while the vehicle and node selection decoders are executed multiple times to construct the solution.
	}
	\label{fig:framework} 
\end{figure}

\begin{figure*}%[!htbp] 
\centering 
 	\setlength{\belowcaptionskip}{-0.5cm}
	\includegraphics[width=0.88\textwidth, trim={1mm 0mm 1mm 1mm},clip]{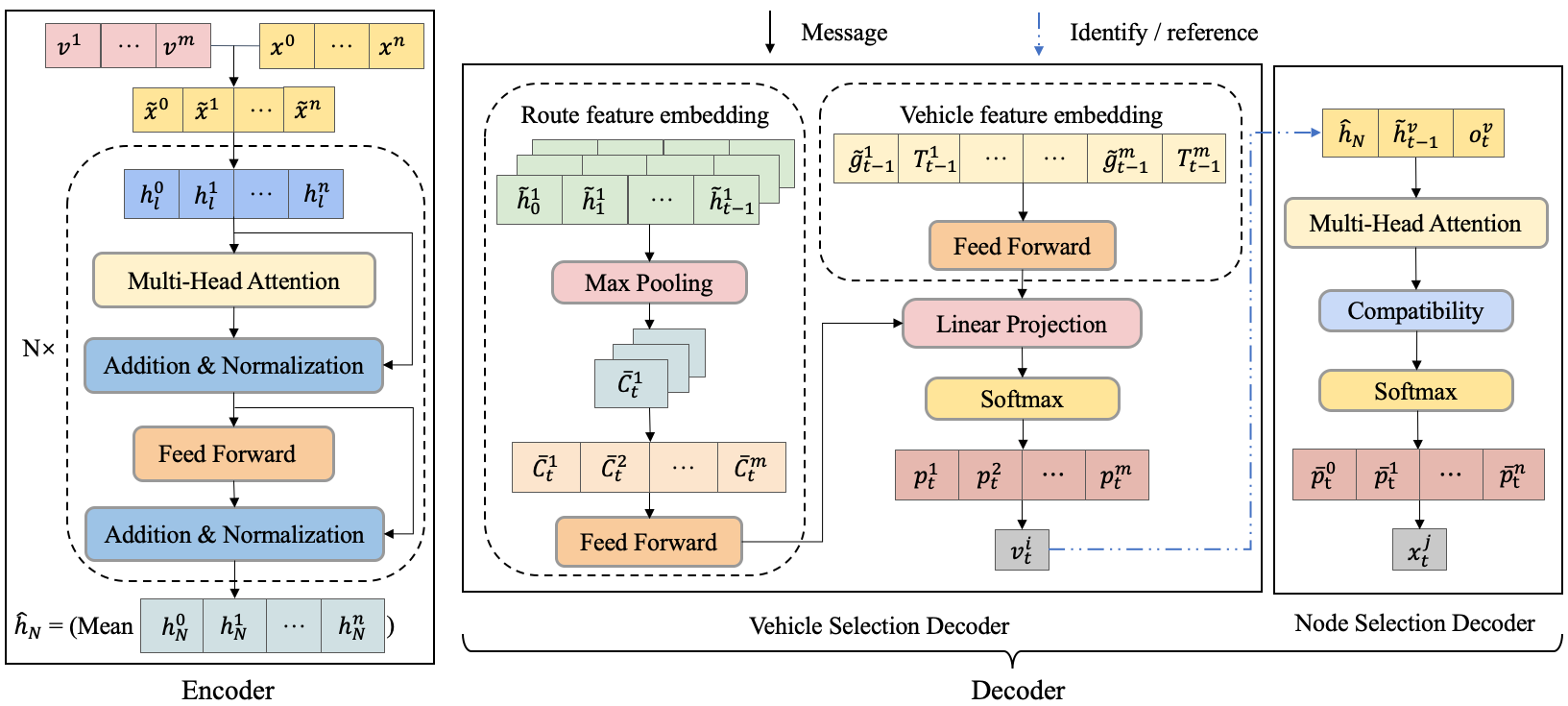} 
	\captionsetup{justification=justified}
	\caption{
	Architecture of our policy network with $m$ heterogeneous vehicles and $n$ customers. It is worth noting that our vehicle selection decoder leverages the vehicle features (last node location and accumulated travel time), the route features (max pooling of the routes for $m$ vehicles), and their combinations to compute the probability of selecting each vehicle.
	} 
	\label{fig:DL_model} 
\end{figure*}

\subsection{Framework of Our Policy Network}

In our approach, we focus on learning a stochastic policy $\pi_{\theta}(a_t|s_t)$ represented by a deep neural network with trainable parameter $\theta$. Starting from the initial state $s_0$, i.e., an empty solution, we follow the policy $\pi_{\theta}$ to construct the solution by complying with the MDP in section~\ref{RL_form} until the terminate state $s_\mathcal{T}$ is reached, i.e., all customers are served by the whole fleet of vehicles. The $\mathcal{T}$ is possibly longer than $n+1$ due to the fact that sometimes vehicles need to return to the depot for replenishment. Accordingly, the joint probability of this process is factorized based on the chain rule as follows,
\begin{equation}
\hspace{-3mm}
    p(s_{\mathcal{T}}|s_0) = \prod_{t=0}^{{\mathcal{T}}-1} \pi_{\theta}(a_t|s_t) p(s_{t+1}|s_{t}, a_{t}),
\end{equation}
where $p(s_{t+1}|s_{t}, a_{t}) = 1$ always holds since we adopt the deterministic state transition rule.

As illustrated in Fig.~\ref{fig:framework}, our policy network $\pi_{\theta}$ is composed of an encoder, a vehicle selection decoder and a node selection decoder. Since a given problem instance itself remains unchanged throughout the decision process, the encoder is executed only once at the first step ($t\!=\!0$) to simplify the computation, while its outputs could be reused in subsequent steps ($ t\!>\!0 $) for route construction. To solve the instance, with raw features processed by the encoder for better representation, our policy network first selects a vehicle ($v^i$) from the whole fleet via the vehicle selection decoder and identify its index, then selects a node ($x^j$) for this vehicle to visit via the node selection decoder at each route construction step. The selected vehicle and node constitute the action for that step, which is further used to update the states. This process is repeated until all customers are served.

\subsection{Architecture of Our Policy Network}
Originating from the field of natural language processing~\cite{vaswani2017attention}, the Transformer model has been successfully extended to many other domains such as image processing~\cite{li2019entangled,yu2019multimodal}, recommendation systems~\cite{sun2019bert4rec,chen2019behavior} and vehicle routing problems~\cite{kool2018attention,xin2020step} due to its desirable capability to handle sequential data. Rather than the sequential recurrent or convolutional structures, the Transformer mainly hinges on the self-attention mechanism to learn the relations between arbitrary two elements in a sequence, which allows more efficient parallelization and better feature extraction without the limitation of sequence-aligned recurrence. Regarding the general vehicle routing problems, the input is a sequence of customers characterized by locations and demands, and the construction of routes could be deemed as a sequential decision-making, where the Transformer has desirable potential to engender high quality solutions with short computation time. Specially, the Transformer-style models~\cite{kool2018attention,xin2020step} adopt an encoder-decoder structure, where the encoder aims to compute a representation of the input sequence based on the multi-head attention mechanism for better feature extraction and the decoder sequentially outputs a customer at each step based on the problem-related contextual information until all customers are visited. To solve the HCVRP with both min-max and min-sum objectives, we also propose a Transformer-style models as our policy network, which is designed as follows.

\begin{figure*}%[!htbp] 
\centering 
 	\setlength{\belowcaptionskip}{-0.5cm}
	\includegraphics[width=1.0\textwidth, trim={0mm 1mm 1mm 0mm},clip]{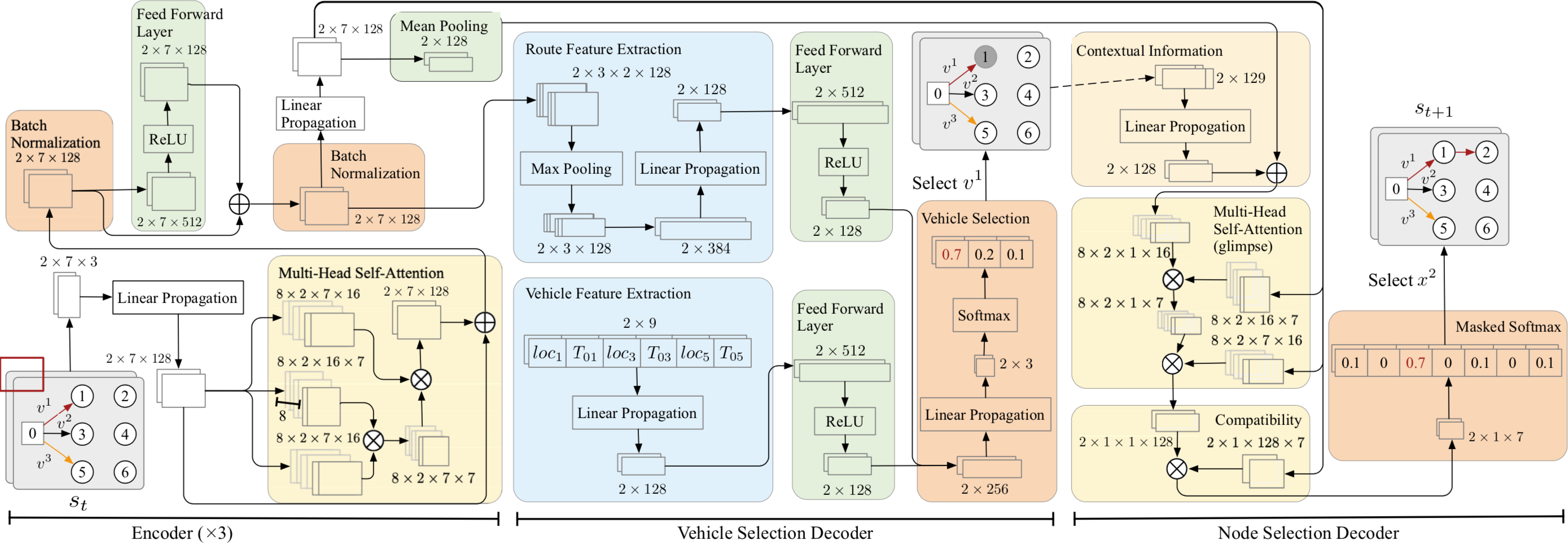} 
	\captionsetup{justification=justified}
	\caption{An illustration of our policy network for two instances with seven nodes and three vehicles, where the red frame indicates the two stacked instances with same data structure. Given the current state $s_t$, the features of nodes and vehicles are processed through the encoder to compute the node embeddings and the graph embedding. In the vehicle selection decoder, the node embeddings of the three tours for three vehicles in current state $s_t$, i.e., $\{\{{h}^0, {h}^1\}, \{{h}^0, {h}^3\}, \{{h}^0, {h}^5\}\}$, are processed for route feature extraction, and the current location and the accumulated travel time of three vehicles are processed for vehicle feature extraction, which are then concatenated to compute the probability of selecting a vehicle. With the selected vehicle $v^1$ in this example, the current node embedding $h^1$ and the current loading ability of this vehicle are first concatenated and linearly propagated, then added with the graph embedding, which are further used to compute the probability of selecting a node with masked softmax, i.e., $\bar{p}^1=\bar{p}^3=\bar{p}^5=0$. With the selected node $x^2$ in this example, the action is represented as $a_t=\{v^1, x^2\}$ and the state is updated and transited to $s_{t+1}$. 
	} 
	\label{fig:detail_model} 
\end{figure*}

As depicted in Fig.~\ref{fig:DL_model}, our policy network adopts an encoder-decoder structure and the decoder consists of two parts, i.e., vehicle selection decoder and node selection decoder. Based on the stipulation that any vehicle has the opportunity to be selected at each step, our policy network is able to search in a more rational and broader action space given the characteristics of HCVRP. Moreover, we enrich the contextual information for the vehicle selection decoder by adding the features extracted from all vehicles and existing (partial) routes. In doing so, it allows the policy network to capture the heterogeneous roles of vehicles, so that decisions would be made more effectively from a global perspective. To better illustrate our method, an example of two instances with seven nodes and three vehicles is presented in Fig.~\ref{fig:detail_model}. Next, we introduce the details of our encoder, vehicle selection decoder, and node selection decoder, respectively.

%\subsection{Encoder}
\textbf{Encoder.}
The encoder embeds the raw features of a problem instance (i.e., customer location, customer demand, and vehicle capacity) into a higher-dimensional space, and then processes them through attention layers for better feature extraction. We normalize the demand $d_0^i$ of customer $x^i$ by dividing the capacity of each vehicle to reflect the differences of vehicles in the heterogeneous fleet, i.e., $\tilde x^i = (s^i, d_0^i / \mathcal{Q}^1, d_0^i / \mathcal{Q}^2, ...d_0^i / \mathcal{Q}^m)$.
Similar to the encoder of Transformer in~\cite{vaswani2017attention,kool2018attention}, the enhanced node feature $\tilde x^i$ is then linearly projected to $h^i_0$  in a high dimensional space with dimension $dim=128$. Afterwards, $h^i_0$ is further transformed to $h^i_N$ through $N$ attention layers for better feature representation, each of which consists of a multi-head attention (MHA) sublayer and a feed-forward (FF) sublayer.

The $l$-th MHA sublayer uses a multi-head self-attention network to process the node embeddings $h_l = ( h^0_l, h^1_l, ..., h^n_l )$. We stipulate that $dim_k=\frac{dim}{Y}$ is the \emph{query/key} dimension,  $dim_v=\frac{dim}{Y}$ is the \emph{value} dimension, and $Y=8$ is the number of heads in the attention.
The $l$-th MHA sublayer first calculates the attention value $Z_{l,y}$ for each head $y\! \in \! \{1,2,...,Y \}$ and then concatenates all these heads and projects them into a new feature space which has the same dimension as the input $h_l$. Concretely, we show these steps as follows,
\begin{equation}
   Q_{l,y} = h_l W^Q_{l,y},\  K_{l,y} = h_l W^K_{l,y},\  V_{l,y} = h_l W^V_{l,y},
   \label{eq:qkv}
\end{equation}
\begin{equation}
   Z_{l,y} \!=\! \text{softmax}(\frac{{Q_{l,y}} {K_{l,y}}^T}{\sqrt{dim_k}})  V_{l,y},
\label{eq:head}
\end{equation}
\begin{equation}
\begin{split}
   \text{MHA} (h_l) &= \text{MHA} (h_l W^Q_l, h_l W^K_l, h_l W^V_l) \\ 
   &=\text{Concat}(Z_{l,1}, Z_{l,2},..., Z_{l,Y}) W^O_l,
\end{split}
\label{eq:atten}
\end{equation}
where $W^Q_{l}, W^K_{l} \in\! \mathbb{R}^{ Y \times dim \times dim_k}$, $W^V_{l} \in \mathbb{R}^{Y \times dim \times dim_v}$, and $W^O_l\in \mathbb{R}^{dim \times dim}$ are trainable parameters in layer $l$ and are independent across different attention layers.

Afterwards, the output of the $l$-th MHA sublayer is fed to the $l$-th FF sublayer with ReLU activation function to get the next embeddings $h_{l+1}$. 
Here, a skip-connection~\cite{he2016deep} and a batch normalization (BN) layer~\cite{ioffe2015batch} are used for both MHA and FF sublayers, which are summarised as follows,
\begin{equation}
   r^i_{l} = BN(h_l^i + \text{MHA}^i(h_l)),
   \label{multi-head}
\end{equation}
\begin{equation}
   h^i_{l+1} = BN(r^i_{l} + FF(r^i_{l})).
   \label{FF}
\end{equation}
Finally, we define the final output of the encoder, i.e., $h_N^i$, as the node embeddings of the problem instance, and the mean of the node embeddings, i.e., $\hat{h}_N=\frac{1}{n} \sum_{i \in X} h_N^i$, as the graph embedding of the problem instance, which will be reused for multiple times in the decoders.

%\subsection{Vehicle Selection Decoder}
\textbf{Vehicle Selection Decoder.} 
Vehicle selection decoder outputs a probability distribution for selecting a particular vehicle, which mainly leverages two embeddings, i.e., \emph{vehicle feature embedding} and \emph{route feature embedding}, respectively.

\emph{1) Vehicle Feature Embedding: } To capture the states of each vehicle at current step, we define the vehicle feature context $C_t^V \in R^{1 \times 3m}$ at step $t$ as follows, 
\begin{equation}
   C_t^V = [\tilde{g}_{t-1}^{1}, T_{t-1}^{1}, \tilde{g}_{t-1}^{2}, T_{t-1}^{2},..., \tilde{g}_{t-1}^{m}, T_{t-1}^{m}],
\end{equation}
where $\tilde{g}_{t-1}^{i}$ denotes the 2-dim location of the last node $g_{t-1}^{i}$ in the partial route of vehicle $v^i$ at step $t-1$, and $T_{t-1}^{i}$ is the accumulated travel time of vehicle $v^i$ till step $t-1$.
Afterwards, the vehicle feature context is linearly projected with trainable parameters $W_1$ and $b_1$ and further processed by a 512-dim FF layer with ReLU activation function to engender the vehicle feature embedding $H_t^V$ at step $t$ as follows, 
\begin{equation}
   H_t^V = FF(W_1 C_t^V + b_1).
\end{equation}

\emph{2) Route Feature Embedding:} 
Route feature embedding extracts information from existing partial routes of all vehicles, which helps the policy network intrinsically learn from the visited nodes in previous steps, instead of simply masking them as did in previous works~\cite{bello2017neural,nazari2018reinforcement,vinyals2015pointer,kool2018attention}.
For each vehicle $v^i$ at step $t$, we define its route feature context $\tilde{C}_t^{i}$ as an arrangement of the node embeddimgs (i.e., $h_N^k$ is the node embeddimgs for node $x^k$), corresponding to the node in its partial route $G_{t-1}^i$. Specifically, the route feature context $\tilde{C}_t^{i}$ for each vehicle $v^i$, $i=1,2,...,m$ is defined as follows,
\begin{equation}
   \tilde{C}_t^{i} = [\tilde{h}_0^{i}, \tilde{h}_1^{i}, ..., \tilde{h}_{t-1}^{i}],
   \label{contexcontex}
\end{equation}
where $\tilde{C}_t^{i} \in R^{t \times dim}$ (the first dimension is of size $t$ since $G_{t-1}^i$ should have $t$ elements at step $t$) and $\tilde{h}_{j}^{i}$ represents the corresponding node embeddings in $h_N$ of the $j$-th node in partial route $G_{t-1}^i$ of vehicle $v^i$. For example, assume $t = 4$ and the partial route of vehicle $v^i$ is $G_{3}^i = \{x^0, x^3,x^3,x^1\}$, then the route feature context of this vehicle at step $t = 4$ would be $\tilde{C}_4^{i} = [ \tilde{h}_0^{i}, \tilde{h}_1^{i}, \tilde{h}_2^{i}, \tilde{h}_3^{i} ] = [ h^0_N, h^3_N, h^3_N, h^1_N ]$.
Afterwards, the route feature context of all vehicles is aggregated by a max-pooling and then concatenated to yield the route context $\hat{C}_t^R$ for the whole fleet, which is further processed by a linear projection with trainable parameters $W_2$ and $b_2$ and a 512-dim FF layer to engender the route feature embedding $H_t^R$ at step $t$ as follows, 
\begin{equation}
    \bar{C}_t^{i} = max(\tilde{C}_t^{i}), i=1,2,...,m,
\end{equation}
\begin{equation}
    \hat{C}_t^R=[\bar{C}_t^{1}, \bar{C}_t^{2}, ..., \bar{C}_t^{m}], 
\end{equation}
\begin{equation}
    H_t^R = FF(W_2 \hat{C}_t^R + b_2).
\end{equation}
Finally, the vehicle feature embedding $H_t^V$ and route feature embedding $H_t^R$ are concatenated and linearly projected with parameter $W_3$ and $b_3$, which is further processed by a softmax function to compute the probability vector as follows, 
\begin{equation}
    H_t = W_3 [H_t^V, H_t^R] + b_3,
\end{equation}
\begin{equation}
    p_t = \text{softmax} (H_t),
    \label{eq:veh_prob}
\end{equation}
where $p_t \in R^m$ and its element $p_t^i$ represents the probability of selecting vehicle $v^i$ at time step $t$. Depending on different strategies, the vehicle can be selected either by retrieving the maximum probability greedily, or sampling according to the vector $p_t$. The selected vehicle $v^i$ is then used as input to the node selection decoder.

%\subsection{Node Selection Decoder}
\textbf{Node Selection Decoder.}
Given node embeddings from the encoder and the selected vehicle $v^i$ from the vehicle selection decoder, the node selection decoder outputs a probability distribution $\bar p_t$ over all unserved nodes (the nodes served in previous steps are masked), which is used to identify a node for the selected vehicle to visit.
Similar to~\cite{kool2018attention}, we first define a context vector $H^c_t$ as follows, and it consists of the graph embedding $\hat{h}_N$, node embedding of the last (previous) node visited by the selected vehicle, and the remaining capacity of this vehicle,
\begin{equation}
    H^c_t = [\hat{h}_N, \tilde{h}_{t-1}^i, o_t^i],
\end{equation}
%where $h_t^{y_v} = h_{\pi_{t-1}^x}^N$ if the vehicle $v$ is selected at $t-1$ step. 
where the second element $\tilde{h}_{t-1}^i$ has the same meaning as the one defined in Eq.~(\ref{contexcontex}) and is replaced with trainable parameters for $t=0$.
The designed context vector highlights the features of the selected vehicle at the current decision step, with the consideration of graph embedding of the instance from the global perspective.
The context vector $H^c_t$ and the node embeddings $h_N$ are then fed into a multi-head attention (MHA) layer to synthesis a new context vector $\hat{H}^c_t$ as a glimpse of the node embeddings~\cite{vinyals2015order}. Different from the self-attention in the encoder, the query of this attention comes from the context vector, while the key/value of the attention come from the node embeddings as shown below, 
\begin{equation}
    \hat{H}_t^c = \text{MHA} (H_t^c W^Q_c, h_N W^K_c, h_N W^V_c),
\end{equation}
where $W^Q_c, W^K_c$ and $W^V_c$ are trainable parameters similar to Eq.~(\ref{eq:atten}). 
%here
We then generate the probability distribution $\bar p_t$ by comparing the relationship between the enhanced context $\hat{H}_t^c$ and the node embeddings $h_N$ through a \emph{Compatibility Layer}. The compatibility of all nodes with context at step $t$ is computed as follows,
\begin{equation}
    u_t = C \cdot tanh(\frac{q_t^T k_t}{\sqrt{dim_k}}),
\end{equation}
where $q_t=\hat{H}^c_t W^Q_{comp}$ and $k_t = h_N W^K_{comp}$ are trainable parameters, and $C$ is set to 10 to control the entropy of $u_t$. Finally, the probability vector is computed in Eq.~(\ref{eq:node_prob})
where all nodes visited in the previous steps are masked for feasibility and element $\bar{p}_t^j$ represents the probability of selecting node $x^j$ served by the selected vehicle $v^i$ at step $t$ as follows,
\begin{equation}
    \bar{p}_t = \text{softmax} (u_t).
    \label{eq:node_prob}
\end{equation}
Similar to the decoding strategy of vehicle selection, the nodes could be selected by always retrieving the maximum $\bar{p}_t^j$, or sampling according to the vector $\bar{p}$ in a less greedy manner.

\IncMargin{0.5em}
\begin{algorithm}[t!]
\small
\SetKwData{Left}{left}\SetKwData{This}{this}\SetKwData{Up}{up}
\SetKwFunction{Union}{Union}\SetKwFunction{FindCompress}{FindCompress}
\SetKwInOut{Input}{input}\SetKwInOut{Output}{output}
\SetKwBlock{DoParallel}{foreach $b=1,2,...,B$ do in parallel}{end}

\Input{Initial parameters $ \theta$ for policy network $\pi_\theta$;\\  initial parameters $ \phi$ for baseline network $v_\phi$; \\
    number of iterations $I$; \\
    iteration size $N$; number of batches $M$;\\
    maximum training steps ${\mathcal{T}}$; significance $\alpha$.
    }
\ForEach{$iter=1,2,...,I$}
{
 Sample $N$ problem instances randomly;\\
%  Reset gradients $d_{\theta} \leftarrow 0$;\\
\ForEach{$i=1,2,...,M$}{
    Retrieve batch $b = N_i$;\\
     \ForEach{$t=0,1,...,{\mathcal{T}}$}
     {
       Pick an action $a_{t,b} \sim \pi_{\theta}(a_{t,b} | s_{t,b})$ ;\\
       Observe reward $r_{t,b}$ and next state $s_{t+1,b}$;\\
     }
  $R_b = -max(\sum\limits_{t=0}^{\mathcal{T}} r_{t,b})$;\\
  GreedyRollout with baseline $v_\phi$ and compute its reward $R^{BL}_b$;\\
$\! d_\theta \!\leftarrow \! \frac{1}{B} \!\sum\limits_{b=1}^B\! (R_b \!- \!R^{BL}_b) \nabla_{\theta} log\  \pi_{\theta}(s_{\mathcal{T},b}  |s_{0,b})$;\\
$\theta \gets \text{Adam}(\theta, d_\theta)$;\\
 }
\If{\textsc{OneSidedPairedTTest}($\pi_{\theta}, v_{\phi}$)  $\textless \alpha$} {$\phi \gets \theta$;
}
}%
\caption{Deep Reinforcement Learning Algorithm}
\label{Algorithm:algorithm}
%\vspace{-2mm}
\end{algorithm}

\subsection{Training Algorithm}
\label{sec:training}
The proposed deep reinforcement learning method is summarized in Algorithm~\ref{Algorithm:algorithm}, where we adopt the policy gradient with baseline to train the policy of vehicle selection and node selection for route construction. The policy gradient are characterized by two networks: 1) the policy network, i.e., the policy network $\pi_{\theta}$ aforementioned, picks an action and generates probability vectors for both vehicles and nodes with respect to this action at each decoding step; 2) the baseline network $v_{\phi}$, a greedy roll-out baseline with a similar structure as the policy network, but computes the reward by always selecting vehicles and nodes with maximum probability. A Monte Carlo method is applied to update the parameters to improve the policy iteratively.
At each iteration, we construct routes for each problem instance and calculate the reward with respect to this solution in line 9, and the parameters of the policy network are updated in line 13. In addition, the expected reward of the baseline network $R_b^{BL}$ comes from a greedy roll-out of the policy in line 10. The parameters of the baseline network will be replaced by the parameters of the latest policy network if the latter significantly outperforms the former according to a paired t-test on several instances in line 15. By updating the two networks, the policy $\pi_{\theta}$ is improved iteratively towards finding higher-quality solutions.

%-------------------------------------------------------------------------------Improvement

% \renewcommand\arraystretch{3}
\begin{table}
\centering
\caption{Parameter Settings of Heuristic Methods.} 
\begin{threeparttable}
\begin{tabular}{c|c}
\toprule
\hline
SISR & \makecell[c]{$\bar{c}=10, L^{max}=10, \alpha=0.01$, $\beta=0.01$,  $T_0=100, T_f=1$,\\  
$ iter=it(size)\tnote{\#},\ it(40)=1.2 \times 10^7, it(60)=1.8 \times 10^7$,\\
$it(80)=2.4 \times 10^7,it(100)=3.0 \times 10^7,
it(120)=3.6 \times 10^7$,\\
$it(140)=4.2 \times 10^7, it(160)=4.8 \times 10^7$} \\
\hline
VNS & \makecell[c]{$r_n=0.1, r_m=0.25$, $p_{ol}=5$,  $p_{ot}=5$, $p_{otd}=1$,\\
$MaxTolerance=0.04$,\\  
$iter=it(size)\tnote{$\ddagger$},\  it(40)=500, it(60)=600$,\\
$it(80)=700, it(100)=800, it(120)=900$,\\
$it(140)=1000, it(160)=1100$} \\
\hline
ACO & \makecell[c]{$m=20$, $\omega_0 = 0.2$, $\alpha=0.1$, $\gamma = 0.1$, \\
$\delta = 3$, $\beta = 3$, $Q_0 = \frac{N-1}{\sum^N_{i=0}\sum^N_{j=0} D_{ij}}$,\\
$iter=it(size)\tnote{$\ddagger$},\  it(40)=10^4, it(60)=1.2 \times 10^4$,\\ 
$it(80)=1.4 \times 10^4, it(100)=1.6 \times 10^4, it(120)=1.8 \times 10^4$,\\
$it(140)=2.0 \times 10^4, it(160)=2.2 \times 10^4$}\\
\hline
FA & \makecell[c]{$\beta  = 1$, $\gamma = 1$, $n=25$, $\alpha = 0.001$,\\
$iter=it(size)\tnote{$\ddagger$}, \ it(40)=1000, it(60)=1200$,\\
$it(80)=1400, it(100)=1600, it(120)=1800$,\\
$it(40)=2000, it(160)=2200$\\
}\\
\hline
\bottomrule
\end{tabular}
\begin{tablenotes}
        \footnotesize
        \item[\#] The iterations are increased as the problem size scales up following the original paper.
        \item[$\ddagger$] The original papers use same iterations for all problem sizes. We linearly increase the iterations as the problem size scales up.
\end{tablenotes}
\end{threeparttable}
\label{parameter}
\end{table}

\section{Computational Experiments}
\label{sec:experiments}

In this section, we conduct experiments to evaluate our DRL method. Particularly, a heterogeneous fleet of fully loaded vehicles with different capacities, start at a depot node and depart to satisfy the demands of all customers by following certain routes, the objective of which is to minimize the longest or total travel time incurred by the vehicles. Moreover, we further verify our method by extending the experiments to benchmark instances from the CVRPLib~\cite{xavier2019cvrplib}. Note that, the HCVRP with min-max and min-sum objectives are both NP-hard problems, and the theoretical computation complexity grows exponentially as problem size scales up.

% \begin{figure}[t] 
% \centering 
% %     \setlength{\abovecaptionskip}{0.2cm}
%  	\setlength{\belowcaptionskip}{-0.5cm}
% 	\includegraphics[width=0.46\textwidth,height=48mm,trim=0 5 0 20]{figs/reward.png} 
% 	\caption{The reward curve of training.}
% 	\label{fig:trend} 
% \end{figure}

\begin{table*}
\caption{DRL Method v.s. Baselines for Three Vehicles (V3).}
\centering
\resizebox{\textwidth}{!}{
\begin{threeparttable}
\begin{tabular}{c|c|c c c|c c c|c c c|c c c|c c c} 
\toprule
& & \multicolumn{3}{c|}{V3-C40}   & \multicolumn{3}{c|}{V3-C60} & \multicolumn{3}{c|}{V3-C80} & \multicolumn{3}{c|}{V3-C100} & \multicolumn{3}{c}{V3-C120}\\
& Method & Obj.   & Gap   & Time   & Obj.   & Gap   & Time  & Obj.   & Gap   & Time & Obj.   & Gap   & Time & Obj.   & Gap   & Time   \\
\midrule
\multirow{9}{*}{\rotatebox{90}{Min-max}} & SISR  & 4.00	& 0\%   & 245s   & 5.58	& 0\%   & 468s	& 7.27	& 0\%   & 752s & 8.89	& 0\%   & 1135s & 10.42 & 0\%   & 1657s \\
& VNS  & 4.17	& 4.25\%   & 115s   &	5.80  & 3.94\%	& 294s	& 7.57	& 4.13\% & 612s & 9.20  & 3.49\%	& 927s &	10.81  & 3.74\%	& 1378s \\
& ACO & 4.31 & 7.75\% & 209s  & 6.18 & 10.75\% & 317s	& 8.14 & 11.97\% & 601s & 10.05 & 13.05\% & 878s & 11.79 & 13.15\% & 1242s \\
& FA  & 4.49 & 12.25\%  & 168s  & 6.30 & 12.90\%	& 285s	& 8.32	& 14.44\% & 397s &	10.11  & 13.72\%	& 522s &	11.98  & 14.97\%	& 667s \\
&AM(Greedy)	& 4.85 & 21.25\%  & 0.37s  & 6.57 & 17.74\%	& 0.54s	& 8.32	& 14.44\% & 0.82s  &	9.98 & 12.26\%	& 1.07s & 11.63  & 11.61\%	& 1.28s\\
&AM(Sample1280) & 4.36 & 9.00\%  & 0.88s  & 5.99 & 7.39\%	&1.19s	& 7.73	& 6.33\% & 1.81s &9.36 & 5.29\%	& 2.51s & 10.94  & 4.99\%	& 3.37s \\
&AM(Sample12800) & 4.31 & 7.75\%  & 1.35s  & 5.92 &6.09\%	&2.46s	& 7.66	& 5.36\% & 3.67s &9.28 & 4.39\%	& 5.17s & 10.85  & 4.13\%	& 6.93s \\
&DRL(Greedy)	& 4.45 & 11.25\%  & 0.70s  & 6.08 & 8.96\%	& 0.82s	& 7.82	& 7.57\% & 1.11s  &	9.42 & 5.96\%	& 1.44s & 10.98  & 5.37\%	& 1.94s\\
&DRL(Sample1280) & 4.17 & 4.25\%  & 1.25s  & 5.77 & 3.41\%	&1.43s	& 7.48	& 2.89\% & 2.25s & 9.07  & 2.02\%	& 3.42s & 10.62 & 1.92\%	& 4.52s \\
&DRL(Sample12800) & 4.14 & 3.50\%  & 1.64s  &5.73 & 2.69\%	& 2.97s	& 7.44	& 2.34\% & 4.56s &9.02  & 1.46\%	& 6.65s &10.56  & 1.34\%	& 8.78s  \\   
\hline
\multirow{10}{*}{\rotatebox{90}{Min-sum}} & Exact-solver & 55.43*	& 0\%  & 71s   & 78.47*  & 0\%	& 214s & 102.42*  & 0\%	& 793s  & 124.61*  & 0\%	& 2512s & -  & - & - \\
& {SISR}  & {{55.79}}	& {{0.65\%}}   & {(254s)}   & {{79.12}}	& {{0.83\%}}   & {(478s)}	& {{103.41}}	& {{0.97\%}}   & {(763s)} & {{126.19}}	& {{1.27\%}}   & {(1140s)}  & {{149.10}}	& {{0\%}}   & {(1667s)}  \\
&{VNS}  & {57.54}	& {3.81\%}   & {109s}   & {81.44}  & {3.78\%}	& {291s}	& {106.18}  & {3.67\%}	& {547s} &{129.32}  & {3.78\%}	& {828s} &{152.56}  & {2.32\%}	& {1217s} \\
&{ACO}  & {60.11} & {8.44\%}  & {196s} & {86.05} & {9.66\%}  & {302s}   & {113.75} & {11.06\%}  & {593s} 	& {140.61} & {12.84\%}  & {859s} & {166.50} & {11.67\%}  & {1189s}   \\
&{FA}  & {59.94} & {8.14\%}  & {164s} & {85.36} & {8.78\%}  & {272s}   & {112.81} & {10.14\%}  & {388s} 	& {138.92} & {11.48\%}  & {518s} & {164.53} & {10.35\%}  & {653s}   \\
&AM(Greedy)	& 66.54 & 20.04\%  & 0.49s  & 91.19 & 16.21\%	& 0.83s	& 117.22	& 14.45\% & 1.01s  &	141.14 & 13.27\%	& 1.23s & 164.57  & 10.38\%	& 1.41s\\
&AM(Sample1280) & 60.95 & 9.96\%  & 0.92s  & 85.74 & 9.26\%	&1.17s	& 111.78	& 9.14\% & 1.79s &135.61 & 8.83\%	& 2.49s & 159.11  & 6.71\%	& 3.30s \\
&AM(Sample12800) & 60.26 & 8.71\%  & 1.35s  & 84.96 & 8,27\%	&2.31s	& 110.94	& 8.32\% & 3.61s &134.72 & 8.11\%	& 5.19s & 158.19  & 6.10\%	& 6.86s \\
&DRL(Greedy)	& 58.99 & 6.42\%  & 0.61s  & 83.06 & 5.85\%	& 1.02s	& 108.44	& 5.88\% & 1.11s  &	131.75 & 5.73\%	& 1.56s & 154.56  & 3.66\%	& 1.96s\\
&DRL(Sample1280) & 57.05 & 2.92\%  & 1.18s  & 80.46 & 2.54\%	& 1.49s	& 105.29	& 2.80\% & 2.34s & 128.63  & 3.23\%	& 3.38s & 151.23  & 1.43\%	& 4.61s \\
&DRL(Sample12800) & 56.84 & 2.54\%  & 1.65s  & 79.92 & 1.85\%	& 2.99s	& 104.63	& 2.16\% & 4.63s & 128.19  & 2.87\%	& 6.74s & 150.73& 1.09\%	& 9.11s  \\
\bottomrule
\end{tabular}
\begin{tablenotes}
        \footnotesize
        % \item[1] {The bold forms indicate that their corresponding algorithms perform best among all learning-based methods. }
        \item[()] {The mark () indicates that the time is computed based on JAVA implementation which is publicly available. For VNS, ACO and FA, we re-implement them in Python since their original codes are not available, where C++ or JAVA was used in the original papers. So the reported time might be different from the ones in the original papers.}
        \item[*] The mark * indicates that all instances are solved optimally.
      \end{tablenotes}
\end{threeparttable}
}
\label{tab:minmax_v3}
\end{table*}

\begin{table*}
\caption{DRL Method v.s. Baselines for Five Vehicles (V5).}
\centering
\resizebox{\textwidth}{!}{
\begin{tabular}{c|c|c c c|c c c|c c c|c c c|c c c
} 
\toprule
& & \multicolumn{3}{c|}{V5-C80}   &
\multicolumn{3}{c|}{V5-C100} &
\multicolumn{3}{c|}{V5-C120} &
\multicolumn{3}{c|}{V5-C140} &
\multicolumn{3}{c}{V5-C160}  \\
& Method & Obj.   & Gap   & Time & Obj.   & Gap   & Time & Obj.   & Gap   & Time  & Obj.   & Gap   & Time  & Obj.   & Gap   & Time   \\
\midrule
\multirow{9}{*}{\rotatebox{90}{Min-max}} & {SISR}  & {{3.90}}	& {{0\%}}   & {(727s)}   & {{4.72}}	& {{0\%}}   & {(1091s)}	& {{5.48}}	& {{0\%}}   & {(1572s)} & {{6.33}}	& {{0\%}}   & {(1863s)} & {{7.16}}	& {{0\%}}   & {(2521s)} \\
&{VNS} & {4.15}	& {6.41\%}   & {725s} & {4.98}	& {7.19\%}   & {1046s} & {5.81}  & {6.02\%}	&  {1454s} & {6.67}	& {5.37\%}   & {2213s}	& {7.53}	& {5.17\%} &  {3321s} \\
&{ACO}  & {4.50} & {15.38\%}  & {612s} & {5.56} & {17.80\%}  & {890s}   & {6.47} & {18.07\%}  & {1285s} 	& {7.52} & {18.80\%}  & {2081s} & {8.51} & {18.85\%}  & {2898s}   \\
&{FA}  & {4.61} & {18.21\%}  & {412s} & {5.62} & {19.07\%}  & {541s}   & {6.58} & {20.07\%}  & {682s} 	& {7.60} & {20.06\%}  & {822s} & {8.64} & {20.67\%}  & {964s}   \\
&AM(Greedy)	& 4.84 & 24.10\%  & 1.08s  & 5.70 & 20.76\%	& 1.31s	& 6.57	& 19.89\% & 1.74s  &	7.49 & 18.33\%	& 1.93s & 8.34  & 16.48\%	& 2.15s\\
&AM(Sample1280) & 4.32 & 10.77\%  & 1.88s  & 5.18 & 8.75\%	&2.64s	& 6.03	& 10.04\% & 3.38s &6.93 & 9.48\%	& 4.47s & 7.75  & 8.24\%	& 5.73s \\
&AM(Sample12800) & 4.25 & 8.97\%  & 3.71s  & 5.11 & 8.26\%	&5.19s	& 5.95	& 8.58\% & 6.94s &6.86 & 8.37\%	& 8.73s & 7.69  & 7.40\%	& 10.69s \\
&DRL(Greedy)	& 4.36 & 11.79\%  & 1.29s  & 5.20 & 10.17\%	& 1.64s	& 5.94	& 8.39\% & 2.38s  &	6.78 & 7.11\%	& 2.43s & 7.61  & 6.28\%	& 3.02s\\
&DRL(Sample1280) & 4.08 & 4.62\%  & 2.66s  & 4.91 & 4.03\%	&3.66s	& 5.66	& 3.28\% & 5.08s &6.51  & 2.84\%	& 6.48s &7.34 & 2.51\%	& 8.52s \\
&DRL(Sample12800) & 4.04 & 3.59\%  & 5.06s  & 4.87 & 3.18\%	& 7.20s	& 5.62	& 2.55\% & 9.65s &6.47  & 2.21\%	& 10.93s &7.30  & 1.96\%	& 13.76s  \\  
% &{DRL($\times$ 8 augment)}  & {\textbf{3.97}} & {\textbf{1.79\%}}  & {40.45s} & {\textbf{4.80}} & {\textbf{1.69\%}}  & {57.58s}  & {\textbf{5.53}} & {\textbf{0.91\%}}  & {77.21s}	& {\textbf{6.37}} & {\textbf{0.63\%}}  & {87.49s} & {\textbf{7.20}} & {{0.56\%}}  & {110.82s}  \\
\hline
\multirow{9}{*}{\rotatebox{90}{Min-sum}} &Exact-solver  & 102.42* & 0\%  & 1787s  & 124.63*  & 0\%	& 6085s	& -  &-	& - & -  &-	& -  & -  &-& -  \\
& {SISR}  & {{103.49}}	& {{1.04\%}}   & {(735s)}   & {{126.35}}	& {{1.38\%}}   & {(1107s)}	& {{149.18}}	& {{0\%}}   & {(1580s)} & {{172.88}}	& {{0\%}}   & {(1881s)} & {{196.51}}	& {{0\%}}   & {(2539s)} \\
&{VNS} & {109.91}	& {7.31\%}   & {538s}   &{133.28}  & {6.94\%}	& {811s}	& {156.37}  & {4.82\%}	& {1386s} &{180.08}  & {4.16\%}	& {2080s} &	{203.95}  &{3.79\%}	& {2896s
}\\
&{ACO}  & {118.58} & {15.78\%}  & {608s} & {146.51} & {17.56\%}  & {865s}   & {171.82} & {15.18\%}  & {1269s} 	& {200.73} & {16.11\%}  & {1922s} & {229.64} & {16.86\%}  & {2803s}   \\
&{FA}  & {116.13} & {13.39\%}  & {401s} & {142.39} & {14.25\%}  & {532s}   & {167.87} & {12.53\%}  & {677s} 	&  {196.48} & {13.65\%}  & {801s} & {223.49} & {13.73\%}  & {955s}   \\
&AM(Greedy)	& 128.31 & 25.28\%  & 0.82s  & 152.91 & 22.69\%	& {1.28s}	& 177.39	& 18.91\% & {1.45s}  &	201.85 & 16.76\%	& {1.69s} & 227.10  & 15.57\%	& {1.81s}\\
&AM(Sample1280) & 119.41 & 16.59\%  & 1.83s  & 144.23 & 15.73\%	&2.66s	& 168.95	& 13.25\% & 3.63s &193.65 & 12.01\%	& 4.68s & 218.67  & 11.28\%	& 5.49s \\
&AM(Sample12800) & 118.04 & 15.25\%  & 3.74s  & 142.79 & 14.57\%	&5.20s	& 167.45	& 12.25\% & 7.02s &192.13 & 11.13\%	& 8.93s & 217.14  & 10.50\%	& 11.01s \\
&DRL(Greedy)	& 108.43 & 5.87\%  & 1.26s  & 131.90 & 5.83\%	& 1.73s	& 154.71	& 3.71\% & 2.11s &	178.78 & 3.41\%	& 3.06s & 202.87  & 3.24\%	& 3.60s\\
&DRL(Sample1280) & 105.54 & 3.05\%  & 2.70s  & 128.63 & 3.21\%	& 4.15s	& 151.39	& 1.48\% & 5.37s & 175.29  & 1.39\%	& 6.83s & 199.16  & 1.35\%	& 8.68s \\
&DRL(Sample12800) & {104.88} & 2.40\%  & 5.38s  & {128.17} & 2.84\%	& 7.61s	& {150.86}	& 1.26\% & 9.24s & {174.80}  & 1.11\%	& 11.30s & {198.66} & 1.09\%	& 13.83s  \\
\bottomrule
\end{tabular}}
\label{tab:minmax_v5}
\vspace{-1mm}
\end{table*}

\subsection{Experiment Settings for HCVRP}
\label{sec:exp_setting}

We describe the settings and the data generation method (which we mainly follow the classic ways in~\cite{bello2017neural,nazari2018reinforcement,vinyals2015pointer,kool2018attention}) for our experiments. Pertaining to MM-HCVRP, the coordinates of depot and customers are randomly sampled within the unit square $[0,1] \times [0,1]$ using the uniform distribution. The demands of customers are discrete numbers randomly chosen from set $\{ 1,2,..,9 \}$ (demand of depot is 0).
To comprehensively verify the performance, we consider two settings of heterogeneous fleets. The first fleet considers three heterogeneous vehicles (named V3), the capacity of which are set to 20, 25, and 30, respectively. The second fleet considers five heterogeneous vehicles (named V5), the capacity of which are set to 20, 25, 30, 35, and 40, respectively. Our method is evaluated with different customer sizes for the two fleets, where we consider 40, 60, 80, 100 and 120 for V3; and 80, 100, 120, 140 and 160 for V5. In MM-HCVRP, we set the vehicle speed $f$ for all vehicles to be 1.0 for simplification. However, our method is capable of coping with different speeds which is verified in MS-HCVRP. 
Pertaining to MS-HCVRP, most of settings are the same as the MM-HCVRP except for the vehicle speeds, which are inversely proportional to their capacities. In doing so, it avoids that only vehicle with the largest capacity is selected to serve all customers to minimize total travel time. Particularly, the speeds are set to $\frac{1}{4}$, $\frac{1}{5}$, and $\frac{1}{6}$ for V3, and $\frac{1}{4}$, $\frac{1}{5}$, $\frac{1}{6}$, $\frac{1}{7}$, and $\frac{1}{8}$ for V5, respectively. 

The hyperparameters are shared to train the policy for all problem sizes. Similar to~\cite{kool2018attention}, the training instances are randomly generated on the fly with iteration size of 1,280,000 and are split into 2500 batches for each iteration. Pertaining to the number of iterations, normally more iterations lead to better performance. However, after training with an amount of iterations, if the improvement in the performance is not much significant, we could stop the training before full convergence, which still could deliver competitive performance although not the best. 
For example, regarding the model of V5-C160 (5 vehicles and 160 customers) with min-max objective trained for 50 iterations, 5 more iterations can only reduce the gap by less than 0.03\%, then we will stop the training. In our experiments, we use 50 iterations for all problem sizes to demonstrate the effectiveness of our method, while more iterations could be adopted for better performance in practice.
The features of nodes and vehicles are embedded into a 128-dimensional space before fed into the vehicle selection and node selection decoder, and we set the dimension of hidden layers in decoder to be 128~\cite{bello2017neural,nazari2018reinforcement,kool2018attention}. In addition, Adam optimizer is employed to train the policy parameters, with initial learning rate $10^{-4}$ and decaying 0.995 per iteration for convergence. The norm of all gradient vectors are clipped to be within 3.0 and $\alpha$ in Section~\ref{sec:training} is set to 0.05.
Each iteration consumes average training time of 31.61m (minutes), 70.52m (with single 2080Ti GPU), 93.02m, 143.62m (with two GPUs) and 170.14m (with three GPUs) for problem size of 40, 60, 80, 100 and 120 regarding V3, and 105.25m, 135.49m, (with two GPUs) 189.15m, 264.45m and 346.52m (with three GPUs) for problem size of 80, 100, 120, 140 and 160 regarding V5. Pertaining to testing, 1,280 instances are randomly generated for each problem size from the uniform distribution, and are fixed for our method and the baselines. Our DRL code in PyTorch is available.\footnote{\href{https://github.com/Demon0312/HCVRP\_DRL}{https://github.com/Demon0312/HCVRP\_DRL}}

\subsection{Comparison Analysis of HCVRP}
\label{sec:comparison}

For the MM-HCVRP, it is prohibitively time-consuming to find optimal solutions, especially for large problem size. Therefore, we adopt a variety of improved classical heuristic methods as baselines, which include: 1) Slack Induction by String Removals (SISR)~\cite{christiaens2020slack}, a state-of-the-art heuristic method for CVRP and its variants; 2) Variable Neighborhood Search (VNS), a efficient heuristic method for solving the consistent VRP~\cite{xu2018variable}; 3) Ant Colony Optimization (ACO), an improved version of ant colony system for solving HCVRP with time windows~\cite{palma2019two}, where we run the solution construction for all ants in parallel to reduce computation time; 4) Firefly Algorithm (FA), an improved version of standard FA method for solving the heterogeneous fixed fleet vehicle routing problem~\cite{matthopoulos2019firefly}; 5) the state-of-the-art DRL based attention model (AM)~\cite{kool2018attention}, learning a policy of node selection to construct a solution for TSP and CVRP. We adapt the objectives and relevant settings of all baselines so that they share the same one with MM-HCVRP. 
We have fine tuned the parameters of the conventional heuristic methods using the grid search~\cite{brito2005grid} for adjustable parameters in their original works, such as the number of shifted points in shaking process, the discounting rate of the pheromones and the scale of the population, and report the best ones in Table~\ref{parameter}. Regarding the iterations, we linearly increase the original ones for VNS, ACO and FA as the problem size scales up for better performance, while the original settings adopt identical iterations across all problem sizes. For SISR, we follow its original setting, where the iterations are increased as the problem size grows up.
To fairly compare with AM, we tentatively leverage two external strategies to select a vehicle at each decoding step for AM, i.e., \emph{by turns} and \emph{randomly}, since it did not cope with vehicle selection originally. The results indicate that vehicle selection \emph{by turns} is better for AM, which is thereby adopted for both min-max and min-sum objectives in our experiments.
Note that, we do not compare with OR-Tools as there is no built-in library or function that can directly solve MM-HCVRP. Moreover, we do not compare with Gurobi or CPLEX either, as our experience shows that they consume days to optimally solve a MM-HCVRP instance even with 3 vehicles and 15 customers. For the MS-HCVRP, with the same three heuristic baselines and AM as used for the MM-HCVRP, we additionally adopted a generic exact solver for solving vehicle routing problems with min-sum objective~\cite{pessoa2020generic}.
The baselines including VNS, ACO, and FA are implemented in Python. For SISR, we adopt a publicly available version\footnote{\href{https://github.com/chenmingxiang110/tsp\_solver}{https://github.com/chenmingxiang110/tsp\_solver}} implemented in JAVA. Note that, the running efficiency of the same algorithm implemented in C++, JAVA, and Python could be considerably different, which will also be analyzed later for running time comparison\footnote{A program implemented in C/C++ might be 20-50 times faster than that of Python, especially for large-scale problem instances. The running efficiency of Java could be comparable to C/C++ with highly optimized coding but could be slightly slower in general.}. All these baselines are executed on CPU servers equipped with the Intel i9-10940X CPU at 3.30 GHz. For those which consume much longer running time, we deploy them on multiple identical servers.

% Before comparing with others, we first evaluate the convergence of our method, which is illustrated by the largest sizes of the two fleets in Fig.~\ref{fig:trend}, i.e., V3-C120 and V5-C160, respectively.  Particularly, V3-C120 refers to 3 vehicles and 120 customers (121 nodes including the depot), and the interpretation holds for all expressions of this type. 
% From the curves we see that the rewards increase very fast for the two sizes until converged, which demonstrate that our method performs well in learning converged policies with high rewards. 

Regarding our DRL method and AM, we apply two types of decoders during testing: 1) Greedy, which always selects the vehicle and node with the maximum probability at each decoding step; 2) Sampling, which engenders $\mathcal{N}$ solutions by sampling according to the probability computed in Eq.~(\ref{eq:veh_prob}) and Eq.~(\ref{eq:node_prob}), and then retrieves the best one. We set $\mathcal{N}$ to 1280 and 12800, and term them as Sample1280 and Sample12800, respectively. 
Then we record the performance of our methods and baselines on all sizes of MM-HCVRP and MS-HCVRP instances for both three vehicles and five vehicles in Table~\ref{tab:minmax_v3} and Table~\ref{tab:minmax_v5}, respectively, which include average objective value, gap and computation time of an instance. Given the fact that it is prohibitively time-consuming to optimally solve MM-HCVRP, the gap here is calculated by comparing the objective value of a method with the best one found among all methods.  

\begin{figure}[t] 
\centering 
 	\setlength{\belowcaptionskip}{-0.3cm}
	\includegraphics[width=0.47\textwidth,height=54mm,trim=10 5 10 20]{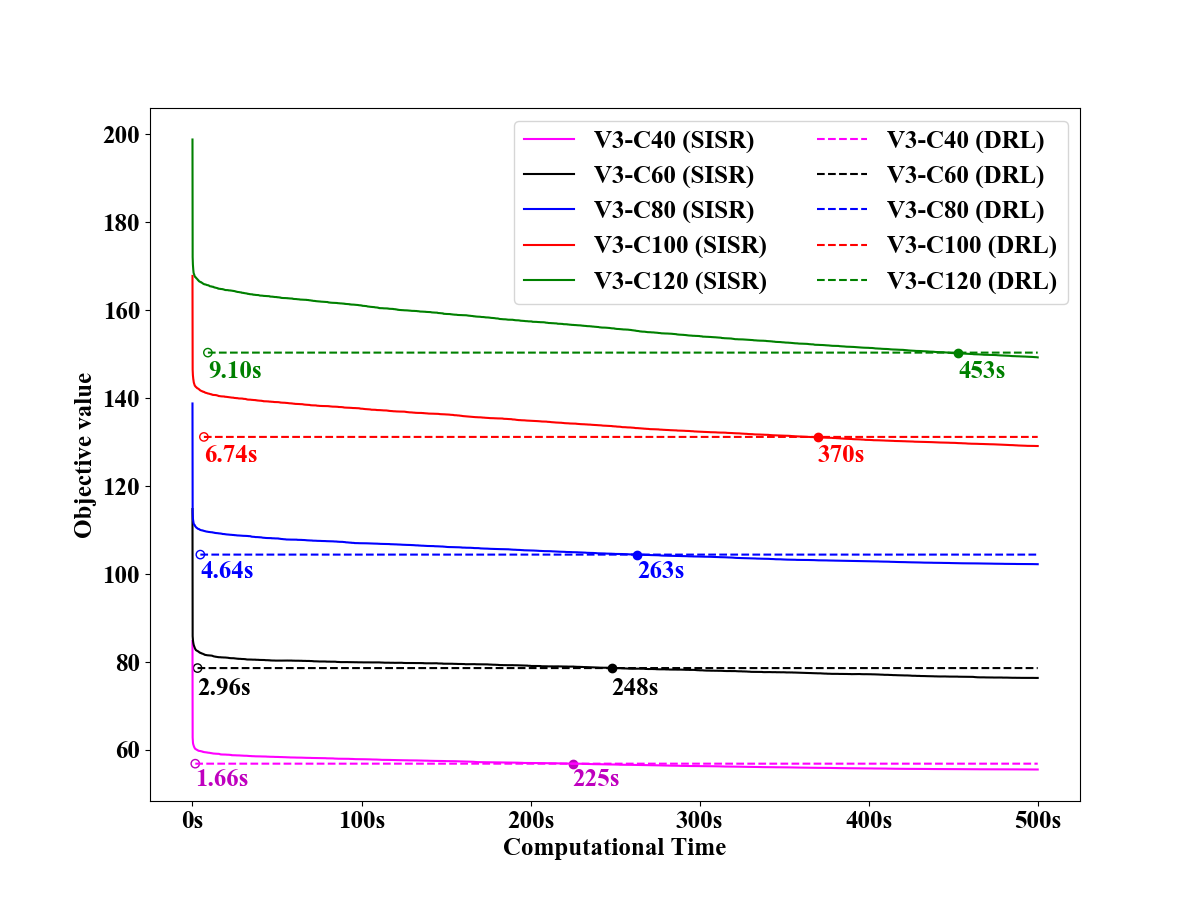}
	\caption{The converge curve of DRL method and SISR (V3).}
	\label{fig:converge_v3} 
\end{figure}

\begin{figure}[t] 
\centering 
 	\setlength{\belowcaptionskip}{-0.3cm}
	\includegraphics[width=0.47\textwidth,height=54mm,trim=10 5 10 20]{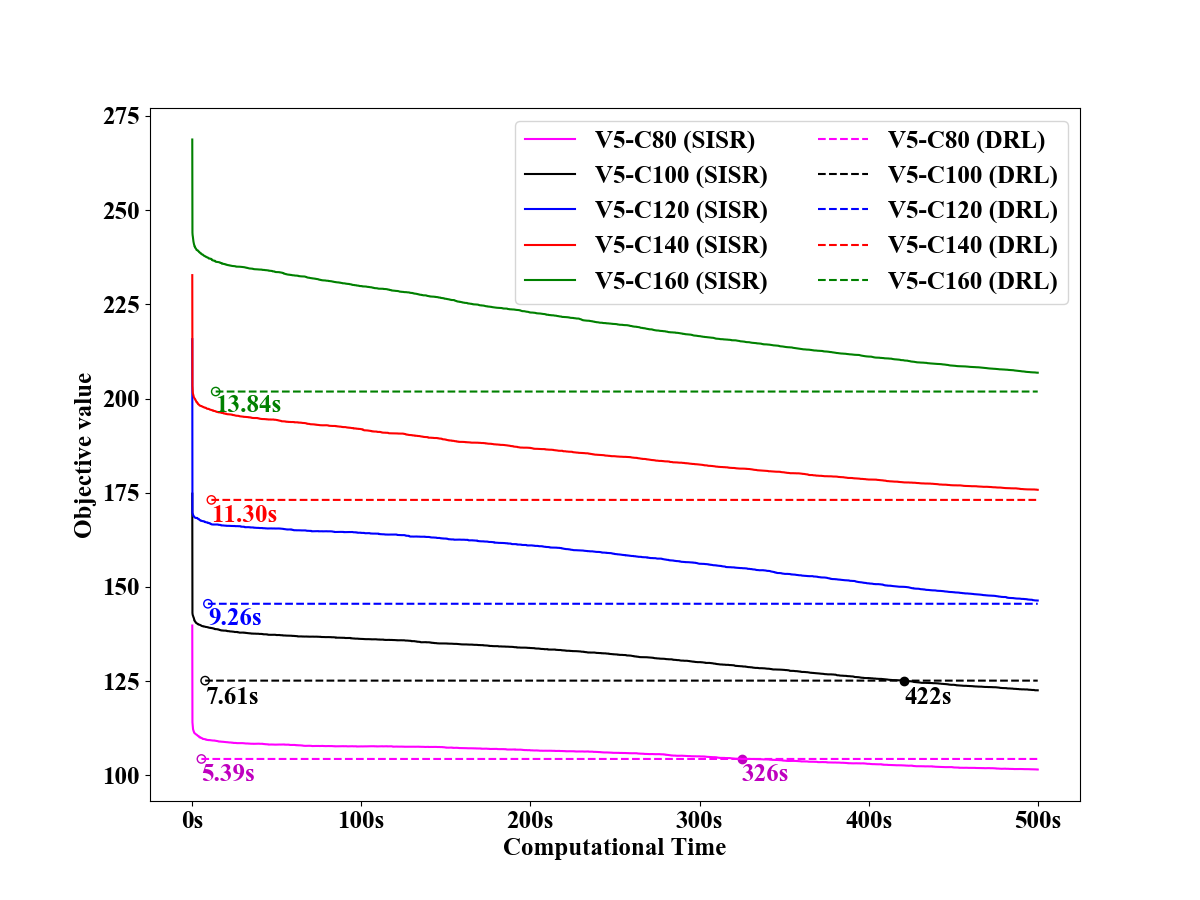}
	\caption{The converge curve of DRL method and SISR (V5).}
	\label{fig:converge_v5} 
\end{figure}

From Table \ref{tab:minmax_v3}, we can observe that for the MS-HCVRP with three vehicles, the Exact-solver achieves the smallest objective value and gap and consumes shorter computation time than heuristic methods on V3-C40 and V3-C60. However, its computation time grows exponentially as the problem size scales up. We did not show the results of the Exact-solver for solving instances with more than 100 customers regarding both three vehicles and five vehicles, which consumes prohibitively long time. Among the three variants of DRL method, our DRL(Greedy) outperforms FA and AM(Greedy) in terms of objective values and gap. Although with slightly longer computation time, both Sample1280 and Sample12800 achieve smaller objective values and gaps than Greedy, which demonstrates the effectiveness of sampling strategy in improving the solution quality. Specifically, our DRL(Sample1280) can outstrip all AM variants and ACO. It also outperforms VNS in most cases except for V3-C40, where our DRL(Sample1280) achieves the same gap with VNS. With Sample12800, our DRL further outperforms VNS in terms of objective value and gap and is slightly inferior to the state-of-the-art heuristic, i.e., SISR and the Exact-solver. Pertaining to the running efficiency, although the computation time of our DRL method is slightly longer than that of AM, it is significantly shorter (at least an order of magnitude faster) than that of conventional methods, even if we eliminate the impact of different programming language via roughly dividing the reported running time by a constant (e.g., 30), especially for large problem sizes. Regarding the MM-HCVRP, similarly, our DRL method outperforms VNS, ACO, FA and all AM variants, which performs slightly inferior to SISR but consumes much shorter running time. Among all heuristic and learning based methods, the state-of-the-art method SISR achieves lowest objective value and gap, however, the computation time of SISR grows almost exponentially as the problem scale increases and our DRL(Sample12800) grows almost linearly, which is more obvious in large-scale problem sizes.

In Table~\ref{tab:minmax_v5}, similar patterns could be observed in comparison with that of three vehicles, where the superiority of DRL(Sample12800) to VNS, ACO, FA and AM becomes more obvious. Meanwhile, our DRL method is still competitive to the state-of-the-art method, i.e., SISR, on larger problem sizes in comparison with Table~\ref{tab:minmax_v3}.
Combining both Tables, our DRL method with Sample12800 achieves better overall performance than conventional heuristics and AM on both the MM-HCVRP and MS-HCVRP, and also performs competitively well 
against SISR, with satisfactory computation time.

To further investigate the efficiency of our DRL method against SISR, we evaluate their performance with a bounded time budget, i.e., 500 seconds. It is much longer than the computation time of our method, given that our DRL computes a solution in a construction fashion rather than the improvement fashion as in SISR. 
In Fig.~\ref{fig:converge_v3}, we record the performance of our DRL method and SISR for MS-HCVRP with three vehicles on the same instances in Table~\ref{tab:minmax_v3} and Table \ref{tab:minmax_v5}, where the horizontal coordinate refers to the computation time and the vertical one refers to the objective value. We depict the computation time of our DRL method using hollow circle, and then horizontally extend it for better comparison. We plot the curve of SISR over time since it improves an initial yet complete solution iteratively. We also record the time when SISR achieves same objective value as our method using filled circle. We can observe that SISR needs longer computation time to catch up the DRL method as the problems size scales up. When the computation time reaches 500 seconds for SISR, our DRL method achieves only slightly inferior objective values with much shorter computation time. For example, the DRL method only needs 9.1 seconds for solving a V3-C120 instance, while SISR needs about 453 seconds to achieve the same objective value. In Fig.~\ref{fig:converge_v5}, we record the results of our DRL method and SISR for five vehicles also with 500 seconds. Similar patterns could be observed to that of three vehicles, where the superiority of our DRL method is more obvious, especially for large-scale problem sizes. For example, on V5-C140 and V5-C160, our DRL method with 11.3 and 13.84 seconds even outperforms the SISR with 500 seconds, respectively. Combining all results above, we can conclude that with a relatively short time limit, our DRL method tends to achieve better performance than the state-of-the-art method, i.e., SISR, and the superiority of our DRL method is more obvious for larger-scale problem sizes. Even with time limit much longer than 500 seconds, our DRL method still achieves competitive performance against SISR.

\begin{figure}[t] 
\centering 
	\setlength{\belowcaptionskip}{-0.4cm}
	\includegraphics[width=0.48\textwidth,height=43mm,trim=5 1 5 20]{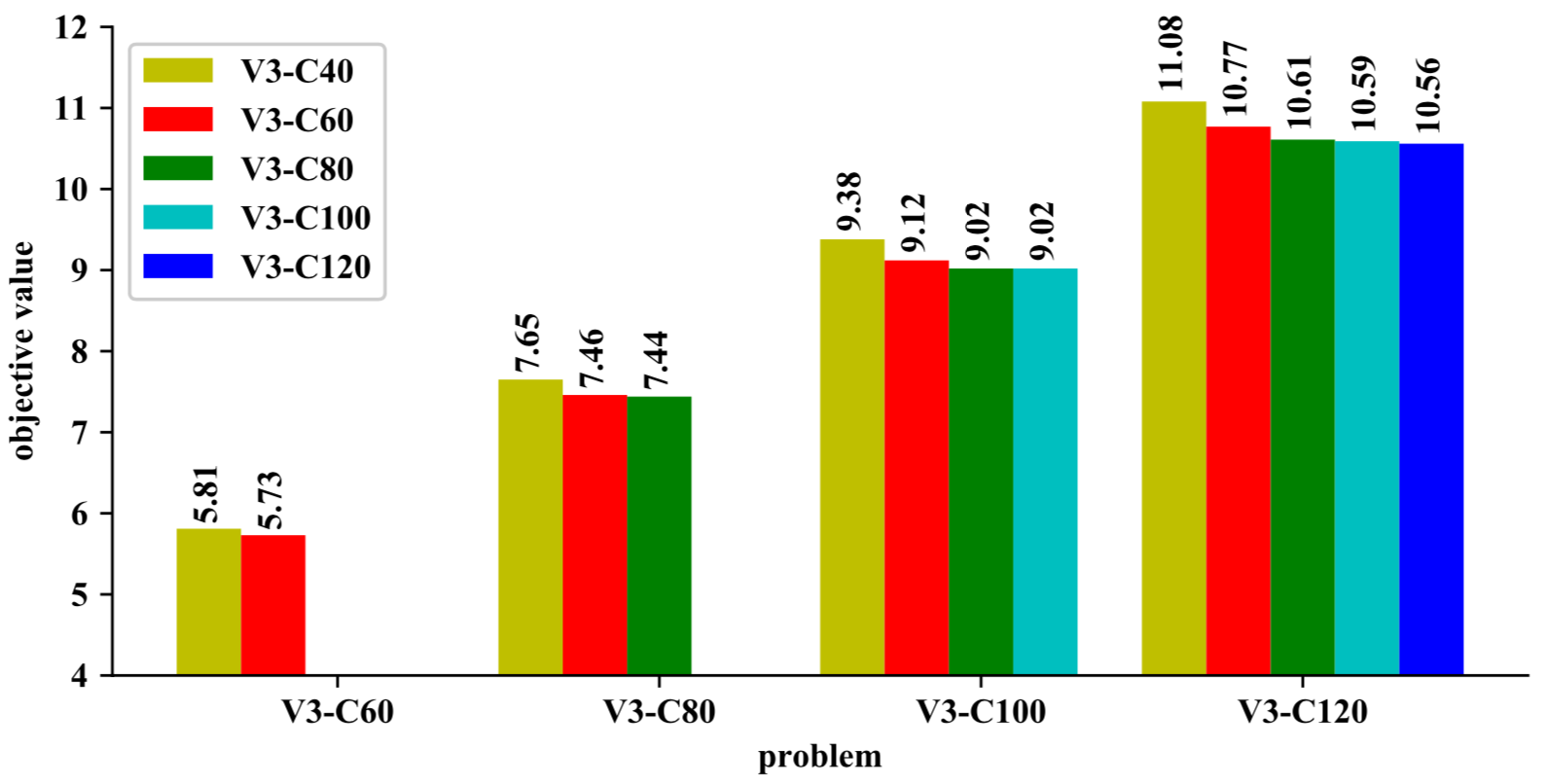} 
	\caption{Generalization performance for MM-HCVRP (V3).}
	\label{fig:comparison_v3} 
\end{figure}

\begin{figure}[t] 
\centering 
	\setlength{\belowcaptionskip}{-0.3cm}
	\includegraphics[width=0.468\textwidth,height=45mm,trim=7 5 3 17]{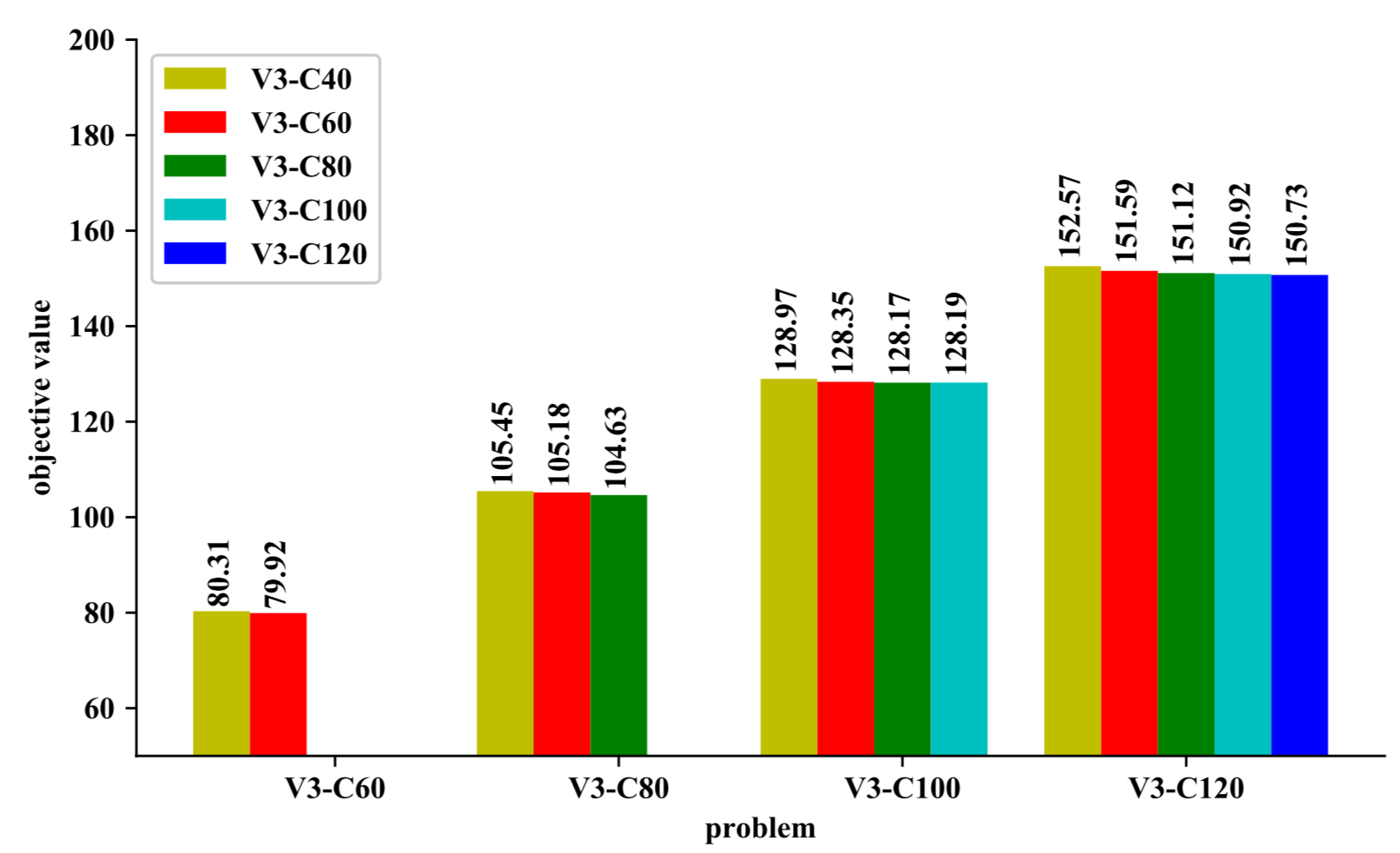}
	\caption{Generalization performance for MS-HCVRP (V3).}
	\label{fig:comparison_minsum_v3} 
\end{figure}

\begin{figure}[t] 
\centering 
 	\setlength{\belowcaptionskip}{-0.3cm}
	\includegraphics[width=0.48\textwidth,height=45mm,trim=5 1 5 20]{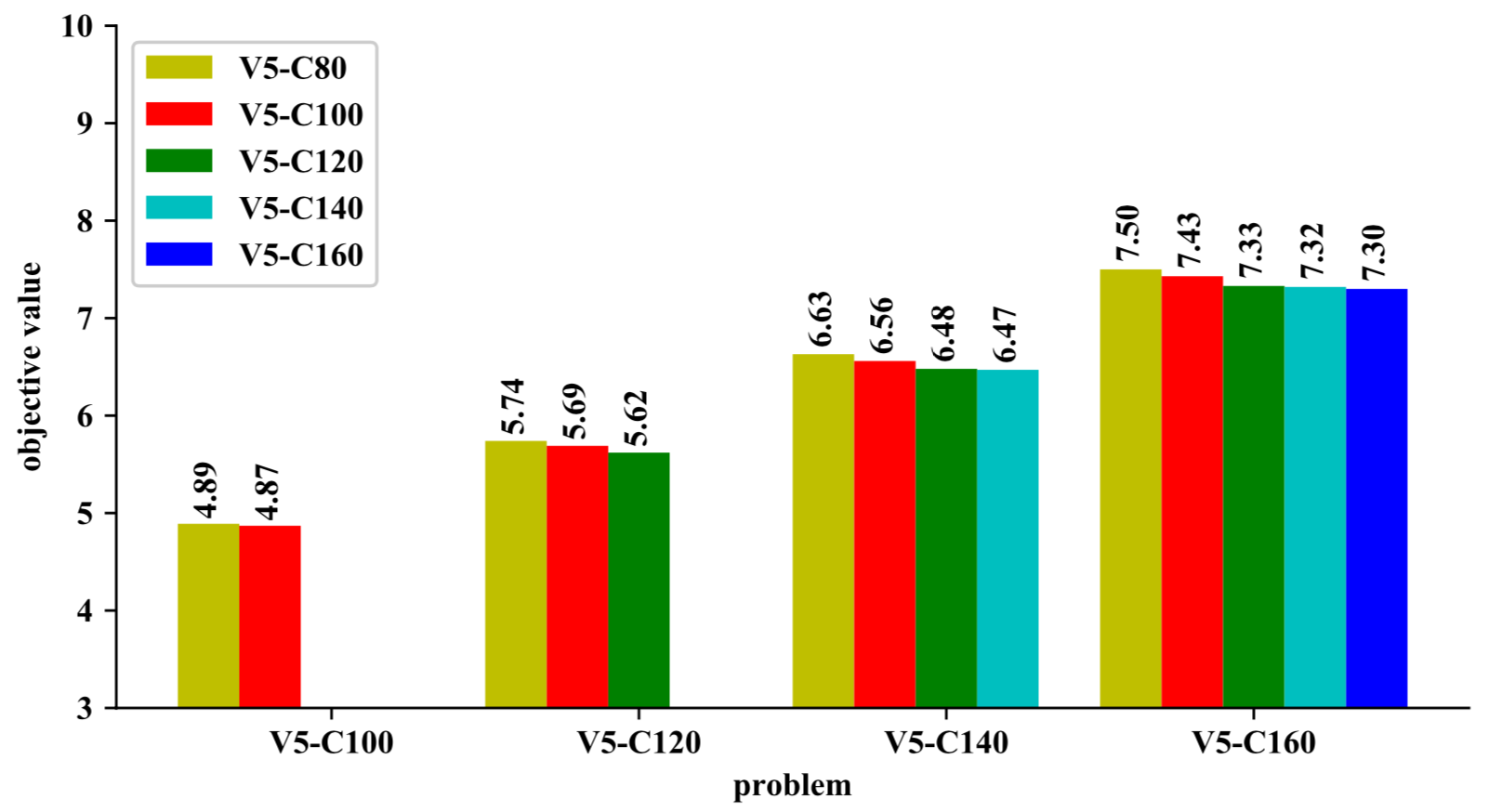} 
	\caption{Generalization performance for MM-HCVRP (V5).}	\label{fig:comparison_v5} 
\end{figure}

\begin{figure}[t] 
\centering 
 	\setlength{\belowcaptionskip}{-0.4cm}
	\includegraphics[width=0.47\textwidth,height=43mm, trim=5 5 5 0]{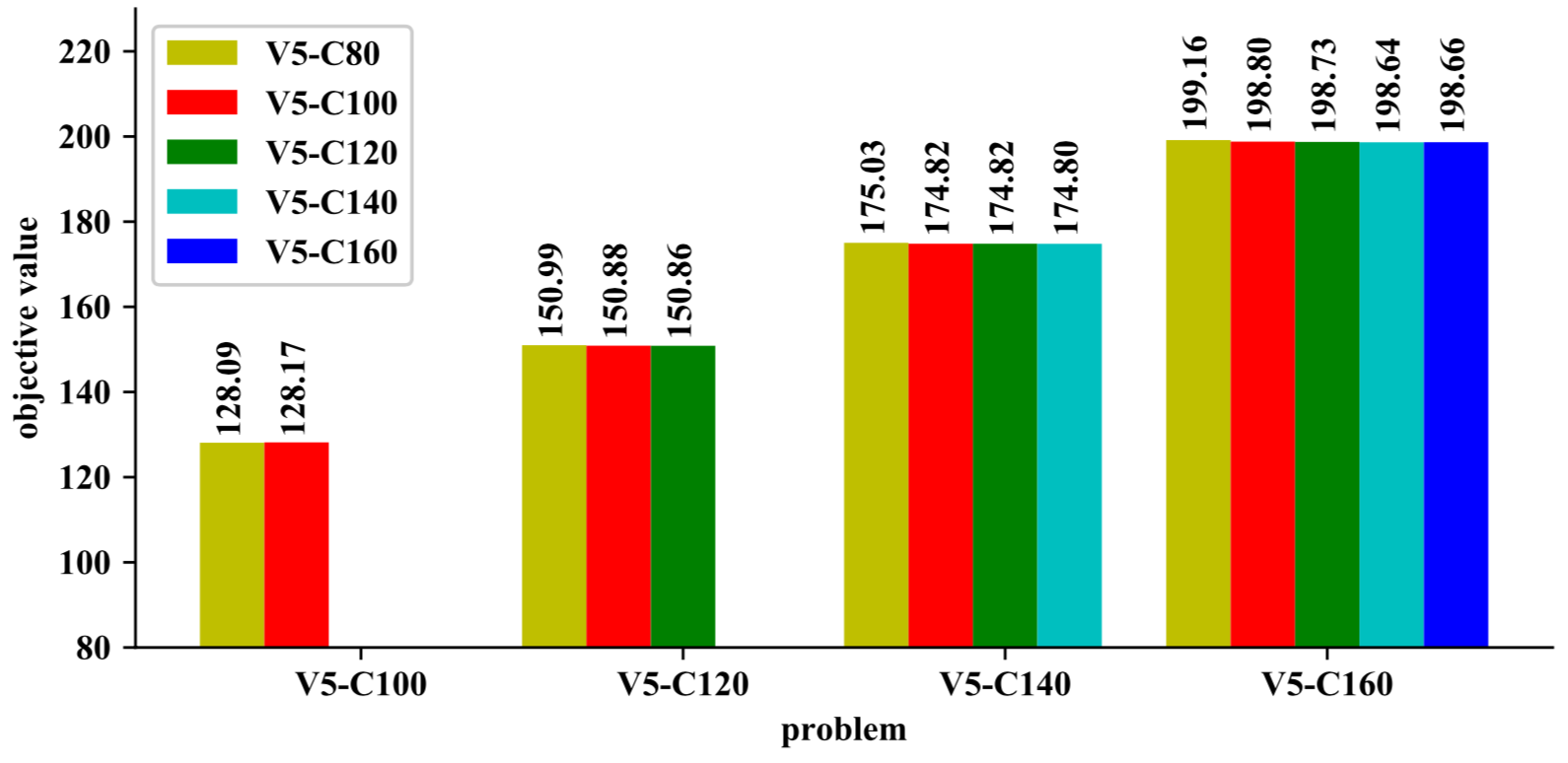} 
	\caption{Generalization performance for MS-HCVRP (V5).}
	\label{fig:comparison_minsum_v5} 
\end{figure}

\subsection{Generalization Analysis of HCVRP}
To verify the generalization of our method, we conduct experiments to apply the policy learnt for a customer size to larger ones since generalizing to larger customer sizes is more meaningful in real-life situations, where we mainly focus on Sample12800 in our method.

In Fig.~\ref{fig:comparison_v3}, we record the results of the MM-HCVRP for the fleet of three vehicles, where the horizontal coordinate refers to the problems to be solved, and the vertical one refers to the average objective values of different methods. We observe that for each customer size, the corresponding policy achieves the smallest objective values in comparison with those learnt for other sizes. However, they still outperform AM and those classical heuristic methods except for V3-C40 in solving problem sizes larger than or equal to 80, where V3-C40 is comparable with the best performed baseline (i.e., VNS), if we refer to Table~\ref{tab:minmax_v3}. Moreover, we also notice that most of the policies learnt for proximal customer sizes tend to perform better than that of more different ones, e.g., the policies for V3-C80 and V3-C100 perform better than that of V3-C40 and V3-C60 in solving V3-C120. The rationales behind this observation might be that proximal customers sizes may lead to similar distributions of customer locations. In Fig.~\ref{fig:comparison_minsum_v3}, we record the results of the MS-HCVRP for the fleet of three vehicles, where similar patterns could be found in comparison with that of the MM-HCVRP, and the policies learnt for other customer sizes outperform all the classical heuristic methods and AM. 

In Fig.~\ref{fig:comparison_v5} and Fig.~\ref{fig:comparison_minsum_v5}, we record the generalization performance of the MM-HCVRP and MS-HCVRP for five vehicles, respectively. Similar patterns to the three vehicles could be observed, where for each customer size, the corresponding policy achieves smaller objective value in comparison with those learnt for other sizes. However, they still outperform most classical heuristic methods and AM in all cases, and are only slightly inferior to the corresponding policies.

\begin{table}
\caption{Our Method v.s. Baselines on CVRPLib.}
\centering
\begin{threeparttable}
\setlength{\tabcolsep}{1.8mm}{
\begin{tabular}{c|c|c|c|c|c|c} 
\toprule
& Dist.  & Instance  & Opt. & Ours & VNS & SISR \\
\midrule
\multirow{10}{*}{\rotatebox{90}{Min-max}} &\multirow{5}{*}{\rotatebox{90}{Uniform}} & P-n60-k10  & - & 306 & {308} & {{293}} \\
&  & A-n61-k9   & - & 319 & {{307}} & {{299}}  \\
&  & E-n76-k7  & - & 372 & {375} & {{362}} \\
&  & A-n80-k10  & - & 795 & {813} & {{776}}  \\
&  & E-n101-k8  & - & 446 & {{455}} & {{428}}  \\
&  & {Avg. Gap} & {-} & {4.11\%} & {4.49\%} & {0\%}  \\
\cline{2-7}
&\multirow{5}{*}{\rotatebox{90}{Non-Uniform}}  & B-n41-k6  & - & 385 & {{371}} & {{359}}  \\
&  & B-n51-k7 & - & 392 & {{378}} & {{369}}  \\
&  & B-n63-k10 & - & 564 & {{558}} & {{540}} \\
&  & M-n101-k10 & - & 419 & {{401}} & {{391}}  \\
&  & CMT11\tnote{1}  & - & 878 & {{869}} & {{858}}  \\
&  & {Avg. Gap} & {-} & {5.48\%} & {2.59\%} & {0\%}  \\
\hline
\multirow{10}{*}{\rotatebox{90}{Min-sum}} 
&\multirow{5}{*}{\rotatebox{90}{Uniform}} & P-n60-k10  & 4009* & 4045 & {4265} & {{4013}} \\
&  & A-n61-k9  & 3984* & 4041 & {4252}  & {{3995}} \\
&  & E-n76-k7  & 4740* & 5035 & {5222} & {{4947}} \\
&  & A-n80-k10  & 11149* & 11454 & {11466}  & {{11186}} \\
&  & E-n101-k8  & 5653* & 5972 & {6114}  & {{5727}} \\
&  & {Avg. Gap} & {0\%} & {2.08\%} & {5.51\%} & {1.28\%}  \\
\cline{2-7}
&\multirow{5}{*}{\rotatebox{90}{Non-Uniform}}  & B-n41-k6 & 4948* & 5327 & {{5015}} & {{4948}} \\
&  & B-n51-k7 & 5235* & 5434 & {{5363}} & {{5236}} \\
&  & B-n63-k10 & 7706* & 7805 & {7825}  & {{7727}} \\
&  & M-n101-k10 & 5443* & 5707 & {{5687}}  & {{5507}} \\
&  & CMT11\tnote{1} & - & 12526 & {{12183}}  & {{11910}} \\
&  & {Avg. Gap} & {0\%} & {4.55\%} & {2.42\%} & {0.29\%}  \\
\bottomrule
\end{tabular}}
\begin{tablenotes}
        \footnotesize
        \item[1] CMT11 has 1 depot and 120 customers (121 nodes).
      \end{tablenotes}
      \end{threeparttable}
      
\label{tab:benchmark}
\end{table}

\subsection{Discussion}
To comprehensively evaluate the performance of our DRL method, we further apply our trained model to solve the instances randomly selected from the well-known CVRPLib\footnote{CVRPLib (\href{http://vrp.atd-lab.inf.puc-rio.br/index.php/en/}{http://vrp.atd-lab.inf.puc-rio.br/index.php/en/}) is a well-known online benchmark repository of VRP instances used for algorithm comparison in VRP literature. More details of CVRPLib can be found in \cite{xavier2019cvrplib}. In our experiment, we select 10 instances and adapt them to our MM-HCVRP and MS-HCVRP settings by adopting their customer locations and demands.} benchmark, half of which follow uniform distribution regarding the customer locations, and the remaining half do not.

In Table \ref{tab:benchmark}, we record the comparison results on CVRPlib, where the Exact-solver is used to optimally solve the MS-HCVRP. Regarding the DRL method, we directly exploit the trained models as in Table \ref{tab:minmax_v3} and Table \ref{tab:minmax_v5} to solve the CVRPLib instances, where the model with the closest size to the instances is adopted. For example, we use the model trained for V3-C60 to solve B-n63-k10. We select SISR and VNS as baselines for both MM-HCVRP and MS-HCVRP, which perform better than other heuristics in previous experiments. Each reported objective value is averaged over 10 independent runs with different random seeds. Note that it is prohibitively long for the Exact-solver to solve MS-HCVRP with more than 100 customers (i.e., CMT11). 
From Table \ref{tab:benchmark}, we can observe that our DRL method tends to perform better than VNS on uniformly distributed instances, and slightly inferior on the instances of non-uniform distribution for both MM-HCVRP and MS-HCVRP. Although inferior to SISR, our DRL method is able to engender solutions of comparable solution quality with much shorter computation time. For example, SISR consumes 1598 seconds to solve the CMT11, while our DRL method only needs 9.0 seconds. We also notice that, our DRL method tends to perform better on uniform distributed instances than that of non-uniform ones if we refer to the gap between our method and the exact method for MS-HCVRP and the SISR for MM-HCVRP.

This observation about different distributions indeed makes sense especially given that as described in Section~\ref{sec:exp_setting}, the customer locations in all training instances follow the uniform distributions, the setting of which is widely adopted in this line of research (e.g.,~\cite{bello2017neural,nazari2018reinforcement,kool2018attention,chen2019learning,xin2020step}). Since our DRL model is a learning method in nature, it does have favorable potential to deliver superior performance when both the training and testing instances are from the same (or similar) uniform distribution. It also explains why our DRL method outperform most of the conventional heuristic methods in Table~\ref{tab:minmax_v3} and Table~\ref{tab:minmax_v5}. When it comes to non-uniform distribution for testing, this superiority does not necessarily preserve, as indicated by the results in Table \ref{tab:benchmark}. However, it is a fundamental out-of-distribution challenge to all learning methods, including our DRL method. The purpose of Table \ref{tab:benchmark} is to reveal when our DRL method may perform inferior to others. Considering that addressing the out-of-distribution challenge is not in the scope of this paper, we will investigate it in future.

\section{Conclusion and Future Work}
\label{sec:conclusion}
In this paper, we cope with the heterogeneous CVRP for both min-max and min-sum objectives. To solve this problem, we propose a learning based constructive heuristic method, which integrates deep reinforcement learning and attention mechanism to learn a policy for the route construction. In specific, the policy network leverages an encoder, a vehicle selection decoder and a node selection decoder to pick a vehicle and a node for this vehicle at each step. Experimental results show that the overall performance of our method is superior to most of the conventional heuristics and competitive to the state-of-the-art heuristic method, i.e., SISR with much shorter computation time. With comparable computation time, our method also significantly outperforms the other learning based method. Moreover, the proposed method generalizes well to problems with larger number of customers for both MM-HCVRP and MS-HCVRP. 

One major purpose of our work is to nourish the development of deep reinforcement learning (DRL) based methods for solving the vehicle routing problems, which have emerged lately. Following the same setting adopted in this line of works~\cite{bello2017neural, nazari2018reinforcement, kool2018attention, chen2019learning, vinyals2015pointer}, we randomly generate the locations within a square of [0,1] for training and testing. The proposed method works well for HCVRP with both min-max and min-sum objectives, but may perform inferior for other types of VRPs, such as VRP with time window constraint and dynamic customer requests. Taking into account the above concerns and other potential limitations that our method may have, in future, we will consider and study the following aspects, 1) time window constraint, and dynamic customer requests or stochastic traffic conditions; 2) generalization to different number of vehicles; 3) evaluation with other classical or realistic benchmark datasets with instances of different distributions (e.g., http://mistic.heig-vd.ch/taillard/problemes.dir/vrp.dir/vrp.html); and 4) improvement over SISR by integrating with active search \cite{bello2017neural} or other improvement approaches (e.g., \cite{wu2021learning}).

\section{Acknowledgement}
This work is supported by the National Natural Science Foundation of China (Grant No. 61803104, 62102228), and Young Scholar Future Plan of Shandong University (Grant No. 62420089964188).

%\bibliographystyle{named}
%\bibliographystyle{plainnat}
\bibliographystyle{ieeetr}
\bibliography{HCVRP}

\begin{IEEEbiography}[{\includegraphics[width=1.0in, height=1.2in, clip,keepaspectratio]{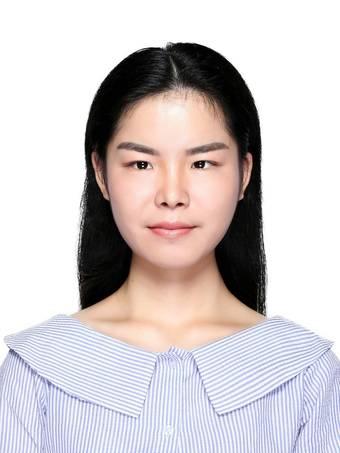}}]{Jingwen Li} received the B.E. degree in computer science from University of Electronic Science and Technology of China, China, in 2018. She is currently pursuing the Ph.D. degree with the department of Industrial Systems Engineering and Management, National University of Singapore (NUS). Her research interests include deep reinforcement learning for combinatorial optimization problems, especially for vehicle routing problems.
\end{IEEEbiography}

\begin{IEEEbiography}[{\includegraphics[width=1.0in, height=1.2in, clip,keepaspectratio]{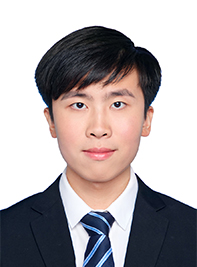}}]{Yining Ma} received the B.E. degree in computer science from South China University of Technology (SCUT), Guangzhou, China, in 2019. He is currently pursuing the Ph.D. degree with the department of Industrial Systems Engineering and Management, National University of Singapore (NUS). His research interests include deep reinforcement learning, combinatorial optimization, and swarm intelligence.
\end{IEEEbiography}

\begin{IEEEbiography}[{\includegraphics[width=1.0in, height=1.2in, clip,keepaspectratio]{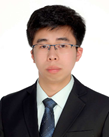}}]{Ruize Gao} received the B.E. degree from Shanghai Jiao Tong University (SJTU), Shanghai, China, in 2020. He is currently pursuing the Ph.D. degree in the Department of Computer Science and Engineering, the Chinese University of Hong Kong (CUHK). His research interests include trustworthy machine learning, especially for adversarial robustness.
\end{IEEEbiography}

\begin{IEEEbiography}[{\includegraphics[width=1in,height=1.25in,clip,keepaspectratio]{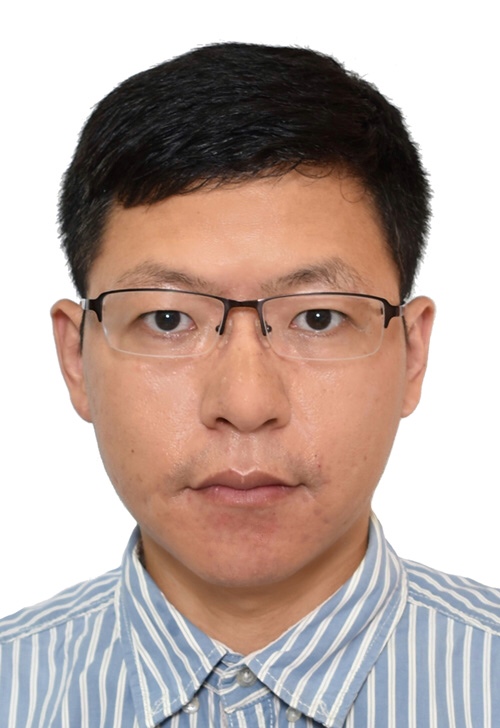}}]{Zhiguang Cao}
received the Ph.D. degree from Interdisciplinary Graduate School, Nanyang Technological University. He received the B.Eng. degree in Automation from Guangdong University of Technology, Guangzhou, China, and the M.Sc. degree in Signal Processing from Nanyang Technological University, Singapore, respectively. He was a Research Assistant Professor with the Department of Industrial Systems Engineering and Management, National University of Singapore, and a Research Fellow with the Future Mobility Research Lab, NTU. He is currently a Scientist with the Singapore Institute of Manufacturing Technology (SIMTech), Agency for Science Technology and Research (A*STAR), Singapore. His research interests currently focus on neural combinatorial optimization.
\end{IEEEbiography}

\begin{IEEEbiography}[{\includegraphics[width=1in,height=1.25in,clip,keepaspectratio]{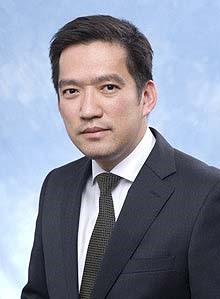}}]{Andrew Lim}
received the Ph.D. degree in computer science in 1992 from the University of Minnesota, Minneapolis. He is currently a Professor with the school of Computing and Artificial Intelligence, Southwest Jiaotong University, China. His works have been published in key journals such as Operations Research and Management Science, and disseminated via international conferences and professional seminars. Before Andrew was recruited by NUS under The National Research Foundation’s Returning Singaporean Scientists Scheme in 2016, he spent more than a decade in Hong Kong where he held professorships in The Hong Kong University of Science and Technology and City University of Hong Kong.
\end{IEEEbiography}

\begin{IEEEbiography}[{\includegraphics[width=1in,height=1.25in,clip,keepaspectratio]{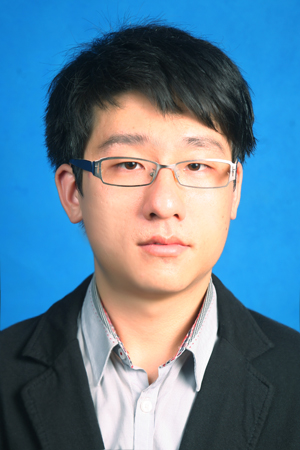}}]{Wen Song}
received the B.S. degree in automation and the M.S. degree in control science and engineering from Shandong University, Jinan, China, in 2011 and 2014, respectively, and the Ph.D. degree in computer science from the Nanyang Technological University, Singapore, in 2018. He was a Research Fellow with the Singtel Cognitive and Artificial Intelligence Lab for Enterprises (SCALE@NTU). He is currently an Associate Research Fellow with the Institute of Marine Science and Technology, Shandong University. His current research interests include artificial intelligence, planning and scheduling, multi-agent systems, and operations research.
\end{IEEEbiography}

\begin{IEEEbiography}[{\includegraphics[width=1in,height=1.25in,clip,keepaspectratio]{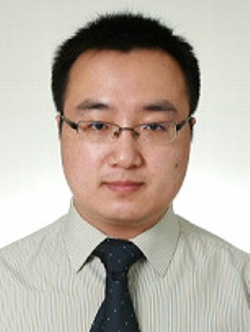}}]{Jie Zhang}
received the Ph.D. degree from the Cheriton School of Computer Science, University of Waterloo, Canada, in 2009. He is currently an Associate Professor with the School of Computer Science and Engineering, Nanyang Technological University, Singapore. He is also an Associate Professor at the Singapore Institute of Manufacturing Technology. During his Ph.D. study, he held the prestigious NSERC Alexander Graham Bell Canada Graduate Scholarship rewarded for top Ph.D. students across Canada. He was also a recipient of the Alumni Gold Medal at the 2009 Convocation Ceremony. The Gold Medal is awarded once a year to honour the top Ph.D. graduate from the University of Waterloo. His papers have been published by top journals and conferences and received several best paper awards. He is also active in serving research communities.
\end{IEEEbiography}

\end{document}